\documentclass{article}

\usepackage{iclr2027_conference,times}

\usepackage[T1]{fontenc}   
\usepackage{textcomp}
\usepackage{amsmath,amssymb,amsfonts,bm}
\usepackage{amsthm}

\usepackage{algorithm}
\usepackage{algpseudocode}
\usepackage{graphicx}
\usepackage{float}      
\usepackage{placeins}
\usepackage{subcaption}
\usepackage{wrapfig}
\usepackage{booktabs}
\usepackage{multirow}
\usepackage{array}
\usepackage{tabularx}
\usepackage{longtable}
\usepackage{xcolor}
\usepackage{tikz}
\usepackage{listings}
\lstdefinestyle{artifact}{ basicstyle=\ttfamily\scriptsize, breaklines=true, breakatwhitespace=false, columns=fullflexible, keepspaces=true, showstringspaces=false, upquote=true, frame=single, rulecolor=\color{gray!45}, xleftmargin=3pt, xrightmargin=3pt, aboveskip=5pt, belowskip=5pt, literate={`}{{\`{}}}1 {\{\{}{{\{\{}}2 {\}\}}{{\}\}}}2 {—}{{-{}-}}2 {–}{{-}}1, }
\usepackage{tcolorbox}
\tcbuselibrary{listings,skins,breakable}
\definecolor{rfbartitle}{HTML}{C4A88A} \definecolor{rfkw}{HTML}{1A5FB4} \definecolor{rfcomment}{HTML}{8A8A8A} \definecolor{rfnum}{HTML}{A03030} \definecolor{rfemph}{HTML}{A0299E} \lstdefinestyle{rfpython}{ language=Python, aboveskip=2pt, belowskip=1pt, basicstyle=\ttfamily\scriptsize, keywordstyle=\color{rfkw}\bfseries, commentstyle=\color{rfcomment}\itshape, stringstyle=\color{teal}, numberstyle=\color{rfnum}, emph={min,max,dim,device,k,keepdim}, emphstyle=\color{rfemph}, morekeywords={Tensor,Dict,Tuple,torch}, breaklines=true, breakatwhitespace=false, columns=fullflexible, keepspaces=true, showstringspaces=false, upquote=true, tabsize=2, literate={-}{{\textendash}}1 {>}{{\textgreater}}1 {<}{{\textless}}1 {φ}{{$\varphi$}}1 {Δ}{{$\Delta$}}1 {—}{{-{}-}}2 {–}{{-}}1, } \lstdefinestyle{rfprompt}{ basicstyle=\ttfamily\scriptsize, aboveskip=2pt, belowskip=1pt, alsoletter={\_}, emph={search\_mode\_guidance,task\_specific\_prompt,search\_mode,max\_turns,write\_by\_turn,task\_id,task\_description,N\_max,max\_samples,max\_attempts,acceptance\_key,metric\_kind,metric\_range,aggregation,reward\_contract,n\_budget,eval\_minutes,rl\_algorithm}, emphstyle=\bfseries\color{rfkw}, breaklines=true, breakatwhitespace=true, breakautoindent=true, breakindent=1.2em, columns=fullflexible, keepspaces=true, showstringspaces=false, upquote=true, frame=none, literate={`}{{\`{}}}1 {\{\{}{{\{\{}}2 {\}\}}{{\}\}}}2 {—}{{-{}-}}2 {–}{{-}}1, } \newtcolorbox{rewardbox}[1]{ enhanced, breakable, colback=white, colframe=black!35, arc=3pt, boxrule=0.5pt, coltitle=black, fonttitle=\small, title={#1}, attach boxed title to top left={xshift=8pt, yshift=-\tcboxedtitleheight/2}, boxed title style={colback=gray!12, colframe=black!35, boxrule=0.4pt, arc=3pt, boxsep=2pt, left=4pt, right=4pt, top=1pt, bottom=1pt}, boxsep=1.5pt, left=4pt, right=4pt, top=8pt, bottom=4pt, before skip=9pt, after skip=6pt, } \definecolor{rfpromptbar}{HTML}{9DC0A5} \newtcolorbox{promptbox}[1]{ enhanced, breakable, colback=rfpromptbar!7, colframe=rfpromptbar!75!black, arc=3pt, boxrule=0.5pt, coltitle=black, fonttitle=\small, title={#1}, attach boxed title to top left={xshift=8pt, yshift=-\tcboxedtitleheight/2}, boxed title style={colback=rfpromptbar!30, colframe=rfpromptbar!75!black, boxrule=0.4pt, arc=3pt, boxsep=2pt, left=4pt, right=4pt, top=1pt, bottom=1pt}, borderline west={1.6pt}{0pt}{rfpromptbar!75!black}, boxsep=1.5pt, left=4pt, right=4pt, top=8pt, bottom=4pt, before skip=9pt, after skip=6pt, } \usetikzlibrary{arrows.meta,positioning,fit,backgrounds,calc}
\definecolor{linkblue}{HTML}{1E5B8C}
\usepackage{url}
\usepackage{hyperref}
\hypersetup{ colorlinks=true, linkcolor=linkblue, citecolor=linkblue, urlcolor=linkblue, breaklinks=true, }
\usepackage[capitalize]{cleveref}

\definecolor{acc}{HTML}{0D9488} \definecolor{obs}{HTML}{2F6B4A}\definecolor{obsbg}{HTML}{E7F3EC} \definecolor{agt}{HTML}{4A3F9E}\definecolor{agtbg}{HTML}{ECEBFB} \definecolor{act}{HTML}{A85D28}\definecolor{actbg}{HTML}{FBEEE2} \definecolor{steel}{HTML}{556072}\definecolor{amber}{HTML}{C2680A} \tikzset{ box/.style ={rectangle, rounded corners=3pt, draw, semithick, align=center, inner sep=4pt, font=\footnotesize}, obsn/.style ={box, draw=obs, fill=obsbg, text=obs}, agtn/.style ={box, draw=agt, fill=agtbg, text=agt}, actn/.style ={box, draw=act, fill=actbg, text=act}, greyn/.style ={box, draw=steel, fill=steel!8, text=steel}, arr/.style ={-{Stealth[length=5pt]}, semithick, draw=steel}, acclbl/.style={font=\scriptsize\itshape, text=acc!70!black}, lbl/.style ={font=\scriptsize\itshape, text=steel}, }

\newcommand{\sys}{\textsc{arbo}}

\newcommand{\coderepo}{https://anonymous.4open.science/r/arbo-30C6/}

\title{Beyond Scripted Search:\\ Sample-Efficient Reward Discovery via Agentic Black-box Optimization}

\author{Minghao Li$^{1}$, Rui Tan$^{1}$ \& Ruihang Wang$^{2}$ \\
$^{1}$Nanyang Technological University \quad $^{2}$Yunnan University \\
\texttt{minghao002@e.ntu.edu.sg}, \texttt{tanrui@ntu.edu.sg}, \texttt{wangruihang@ynu.edu.cn}}

\iclrfinalcopy 

\begin{document}

\maketitle
\ificlrfinal\lhead{}\renewcommand{\headrulewidth}{0pt}\fi

\begin{abstract}
Designing dense reward functions for low-level reinforcement learning (RL) control remains difficult.
Recent work uses large language models (LLMs) to iteratively generate and refine reward functions using policy-training feedback within scripted search algorithms.
However, evaluating each candidate requires a full RL training run, making sample efficiency a central challenge for reward search on complex control tasks.
To address this limitation, we propose an \underline{A}gentic \underline{R}eward \underline{B}lack-box \underline{O}ptimization (\textbf{\sys{}}) framework, in which an LLM agent builds the search strategy at run time from an evaluation history maintained as its persistent workspace.
The evaluation history comprises two components: \textit{observations} maintained by the evaluation oracle, including candidate scores, per-term training curves, and error tracebacks; and an agent-maintained \textit{belief} that records diagnoses and intended next steps.
The agent queries both with tools and generates the next batch of reward candidates, rather than generating them in a single pass from a fixed prompt.
Across four control domains, \sys{} achieves gains of 29.9\% in manipulation success rate and 192.8\% in power-grid score over baseline means under a shared evaluation budget.
Ablations examine each component's contribution and sensitivity to backbone choice.
\end{abstract}

\section{Introduction}
Dense reward functions are critical to reinforcement learning (RL) in low-level control.
The reward is the sole optimization signal for a policy, and a task's own metric, such as success rate, is too sparse to drive learning~\citep{isaacgym,bidexhands}.
Dense RL rewards that measure progress toward the goal enable learning on these tasks. This process requires substantial expert effort to design appropriate reward functions~\citep{ng1999policy}.
Large language models (LLMs) can now automate reward design.
Given the task description and its environment code, an LLM writes a dense, human-readable reward function directly as executable code~\citep{yu2023language,eureka,revolve}.
Since reward quality can be assessed only through the resulting trained policy, these methods refine their proposals iteratively using the evaluation score.
Rewards designed by LLMs have matched or surpassed hand-engineered rewards on these robotic tasks.

\begin{figure}[!htbp]
\centering
\includegraphics[width=\textwidth]{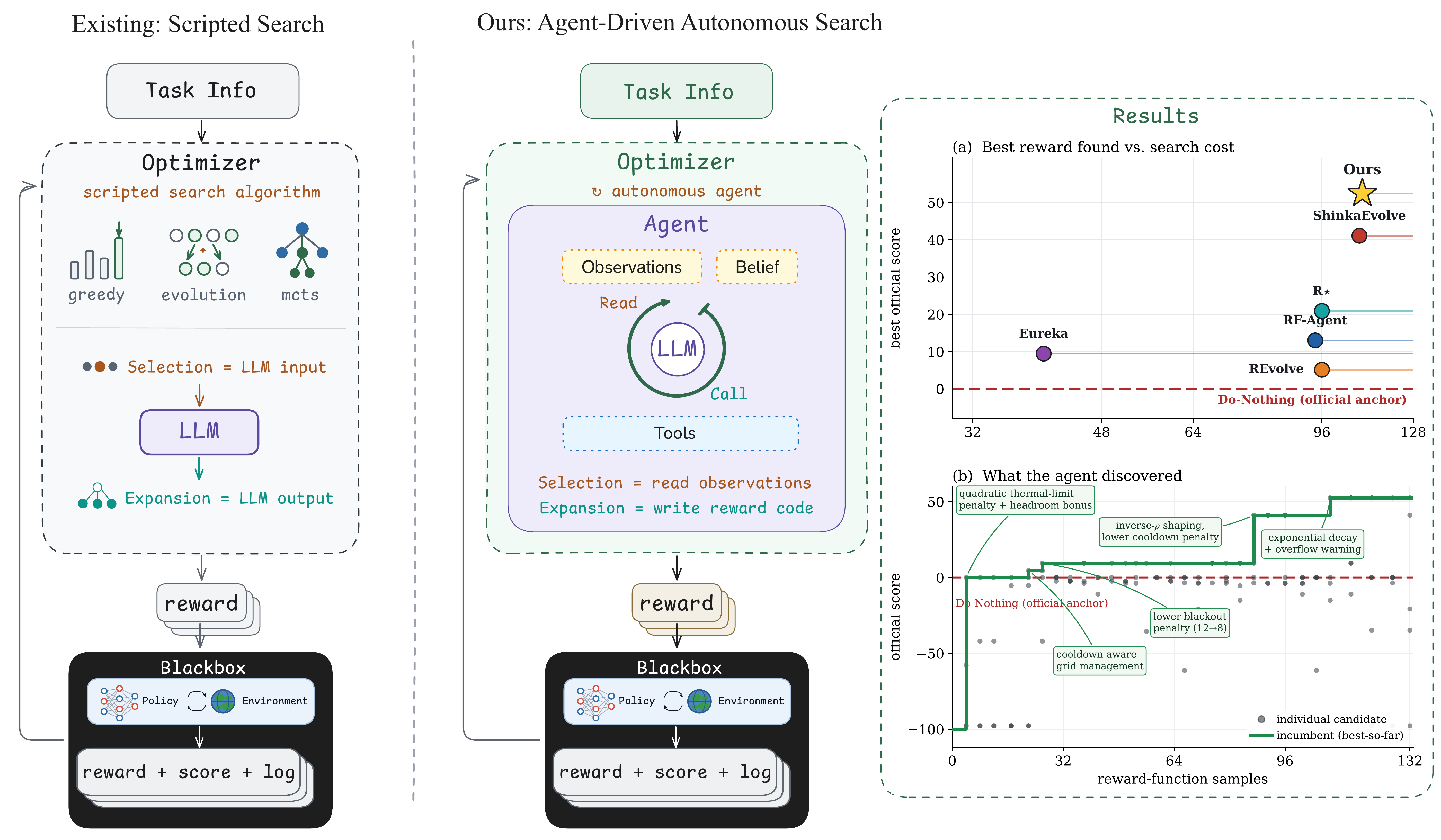}
\vspace{-1.3em}
\caption{\textbf{Agentic Reward Black-box Optimization.} Where prior methods wrap the LLM in a scripted outer search rule, \sys{} lets the agent analyze the history and propose the next candidates.}
\label{fig:overview}
\vspace{-0.5em}
\end{figure}

However, such iterative search is time-consuming, as each candidate reward can be evaluated only by training a policy with it in simulation.
A single evaluation takes hours in control tasks for robotics and cyber-physical systems (CPS)~\citep{isaacgym,grid2op}, and a full search therefore takes hundreds of hours.
We measure the cost of a search by the number of reward candidates it evaluates, as this policy training dominates every other cost in the LLM search loop.
Reward discovery is black-box optimization, yet existing acceleration algorithms for black-box search do not apply to it.
Zeroth-order estimators~\citep{zerothorder} perturb a parameter vector to recover a descent direction, but such perturbations affect only the numerical constants of a reward program, not its structure.
Evolution strategies~\citep{cmaes} adapt a sampling distribution over a continuous search space, but the space of reward programs is discrete.
Surrogate-based optimization~\citep{boreview} maintains a belief over the objective inferred from the observations, but fitting a statistical model here would require many full RL training runs in advance.
Accelerating reward discovery therefore requires improving the sample efficiency of the search strategy.

Existing automated reward design methods can be viewed as black-box optimizers, each defined by how it represents past evaluations and how it uses them to generate subsequent candidates.
Reward discovery methods treat the reward program itself as the search variable, mutating expressions directly~\citep{gpreward} or having an LLM write it as code~\citep{yu2023language}, and these LLM-based methods differ in how they retain history and generate subsequent candidates.
Eureka~\citep{eureka,selfalign} retains only the best candidate from each batch and refines it greedily.
REvolve~\citep{revolve,evonav} maintains a population with associated scores and selects parents by score alone.
RF-Agent~\citep{rfagent} maintains a search tree with value estimates and expands the node with the highest upper confidence bound.
R$\star$~\citep{rstar} tracks the current stage and alternates structural mutation with numerical search under a predefined schedule.
However, all these methods select historical candidates and use manually specified proposal rules, resulting in limited sample efficiency, as \Cref{sec:formulation} demonstrates.
Meta-black-box optimization~\citep{metabox} replaces such scripted rules with a meta-level policy that typically configures predefined operators, but it learns this policy by meta-training over many problem instances, which is unaffordable when every evaluation is a full RL training run.

To address this limitation, we propose \sys{}, an agentic black-box optimization framework for automated reward discovery, in which an LLM agent builds the search strategy at run time from an evaluation history maintained as its persistent workspace.
Prior methods specify in advance both the history they retain and the proposal rule they use, whereas \sys{} retains the complete history and allows the agent to determine at runtime which records to consult and which candidates to generate next, as \Cref{fig:overview} illustrates.
The history comprises two components, the \textit{observations} appended only by the evaluation oracle, which retain each candidate with its per-term curves and traceback, and the \textit{belief} written only by the agent, which records its diagnoses and intent.
This separation is inspired by Bayesian optimization: the agent maintains a revisable interpretation of past observations to guide candidate proposals, without explicitly modeling a probabilistic posterior or optimizing an acquisition function. 
As the proposal operator, the agent queries both components with shell and file operation tools~\citep{react,toolformer} and generates subsequent candidates itself, using evaluation evidence rather than a predefined proposal rule.
The evaluation oracle, the observations, and a fixed exploration schedule remain external to the agent, preserving evaluation as a black box beyond its control.
Our agentic black-box optimization framework adapts the search to each task directly from the evaluations it has already collected.

We evaluate \sys{} on robotic and CPS tasks across four control domains, including dexterous manipulation~\citep{bidexhands,isaacgym}, power-grid operation~\citep{grid2op,l2rpn}, legged locomotion~\citep{isaaclab}, and traffic-signal control~\citep{resco,sumo}, against five automated LLM search methods.
Under the same budget, \sys{} discovers reward functions that improve the mean success rate across the ten Bi-DexHands tasks by $29.9\%$ over the baseline mean.
Power-grid evaluation further confirms that \sys{} outperforms the baselines by $192.8\%$ on average.
Ablations show that the seven ablated variants lose $29.6\%$ of the gain on average.
This loss ranges from $6.0\%$ on traffic-signal control through $19.3\%$ on locomotion and $35.3\%$ on manipulation to $57.7\%$ on the power grid.
These results show that \sys{} outperforms scripted LLM search in efficient reward discovery on robotic and CPS tasks.


\vspace{-0.65\baselineskip}
\section{Related Work}
\label{sec:related}
\vspace{-0.65\baselineskip}
\textbf{Automated reward discovery.} Automated methods replace the expert effort of designing dense rewards~\citep{ng1999policy,pbadvice,dpbrs,rewardmachines,ng2000algorithms,christiano2017deep,envshaping}.
One-shot discovery has an LLM write the reward as code from the task description~\citep{yu2023language}, without iterative refinement using policy training feedback.
Iterative discovery closes this gap by revising candidates against training feedback, from genetic programming~\citep{gpreward} to LLM search that refines the best candidate greedily~\citep{eureka,selfalign}, evolves a population~\citep{revolve,rstar,evonav}, or expands a search tree or graph~\citep{rfagent,regot}. 
Other systems extend this loop with domain randomization~\citep{dreureka}, a relabeled replay buffer shared across candidates~\citep{lares}, or meta RL for adaptive CPS operations~\citep{dcopilot}.
RDA~\citep{rda} aligns rewards with task instructions through vision-language-model judgments of rollouts, RHO~\citep{rho} searches policy code evaluated by execution, and \citet{retrorho} refine an agent's own harness from past trajectories.
These predefined search procedures may limit sample efficiency when each evaluation requires policy training, as examined in \Cref{sec:formulation}.
Efficient reward search therefore requires an optimizer that analyzes the accumulated history itself and proposes high-quality candidates.

\textbf{LLM agents for optimization.} Reward discovery belongs to a broader class of problems in which LLMs drive gradient-free search and every candidate is scored by a black-box evaluation.
Early studies regard the LLM as a variation operator inside a classical evolutionary loop and leave selection and population management to the surrounding algorithm.
ELM~\citep{elm} mutates programs with diff models under a MAP-Elites archive, and LMX~\citep{lmx} performs crossover by prompting with concatenated parents and parsing the completion as offspring.
FunSearch~\citep{funsearch} scales this operator role with an island-based program database that feeds sampled high-scoring programs into each prompt, and AlphaEvolve~\citep{alphaevolve}, together with open reimplementations~\citep{openevolve,codeevolve}, extends the recipe to entire programs and general algorithm discovery.
Later systems refine the loop without changing its shape.
ReEvo~\citep{reevo} pairs the generator LLM with a reflector LLM whose reflections serve as verbal gradients, ShinkaEvolve~\citep{shinka} improves sample efficiency through adaptive parent sampling and novelty-based rejection, and GEPA~\citep{agrawal2025gepa} mutates prompts by reflecting on execution traces under a Pareto-frontier archive.
A parallel, minimalist branch dispenses with explicit evolutionary operators, as OPRO~\citep{opro} prompts the LLM with the sorted trajectory of past solution-score pairs and asks for a better one directly.
To further improve sample efficiency, \citet{sga} introduce differentiable simulations into discrete scientific hypotheses for continuous parameter optimization.
These approaches embed LLM-based generation within externally specified search procedures that govern candidate selection and exploration~\citep{llm4optsurvey}.
Meta-black-box optimization~\citep{metabox} learns such a procedure by training a meta-level policy over a problem distribution to configure a low-level optimizer, whereas \sys{} builds the search strategy at run time on each task, with an LLM agent writing the candidates directly.
The search history lives in the agent's file system, and the agent queries it with tools before writing each batch.

\vspace{-0.45\baselineskip}
\section{Problem Statement}
\label{sec:formulation}
\vspace{-0.45\baselineskip}
In this section, we first define the reward design problem. Second, we formulate reward discovery as black-box optimization and decompose existing algorithms into a history representation and a proposal operator. We then illustrate limitations of fixed search rules on a robotics control task.

\textbf{The reward design problem.}
A control task is a Markov decision process (MDP) $\langle S,A,T,R,\gamma,\rho_0\rangle$ with state space $S$, action space $A$, transition dynamics $T$, discount $\gamma$, and initial-state distribution $\rho_0$.
Of these, the reward function $R$ is the sole component left to the designer.
The reward design problem (RDP)~\citep{sorg2010reward} makes this explicit.
The task supplies a world model $M=(S,A,T)$ with no reward, and choosing $R$ from the space $\mathcal{R}$ of maps $S\times A\to\mathbb{R}$ completes an MDP $(M,R)$.
A fixed learning algorithm $A_{\mathcal{M}}$, the one each benchmark ships, trains under $R$ and returns the best policy reached during training, $\pi=A_{\mathcal{M}}(R)$, and a task metric $F(\pi)$, such as success rate, scores the result.
The evaluation metric is fixed, and the search varies the reward to improve policy performance.
The RDP therefore asks for the reward that maximizes the task metric:
\begin{equation}
  R^\star=\arg\max_{R\in\mathcal{R}} F\big(A_{\mathcal{M}}(R)\big).
  \label{eq:rdp}
\end{equation}

Each candidate is executable reward code emitted by a language model $p_\theta$ conditioned on a task specification $\mathcal{T}$, comprising the environment code and its  evaluation metric, and evaluating it requires training $\pi=A_{\mathcal{M}}(R)$ for a fixed step budget and measuring $F$.
Writing $F(R):=F\big(A_{\mathcal{M}}(R)\big)$ for the induced objective on rewards, the goal is to emit code $R\sim p_\theta(\mathcal{T})$ maximizing $F(R)$.

\textbf{Reward discovery as black-box optimization.}
The objective $F(R)$ of \Cref{eq:rdp} admits no analytic form.
A candidate reward is evaluated by integrating it into the environment, training a policy, and measuring task performance.
Reward discovery is therefore black-box optimization, and every search strategy for it proceeds by the same iteration
\begin{equation}
  R_{t+1}\sim g\big(\cdot\mid\mathcal{T},\,\phi(\mathcal{D}_t)\big),\qquad
  y_{t+1}=F(R_{t+1}),\qquad
  \mathcal{D}_{t+1}=\mathcal{D}_t\cup\{(R_{t+1},y_{t+1})\},
  \label{eq:generic}
\end{equation}
where $t$ counts the queries issued, $\mathcal{D}_t=\{(R_i,y_i)\}_{i\le t}$ is the evaluation history, $\phi$ is the history representation that condenses it into the context the proposer consumes, and $g$ is the proposal operator that turns this context into the next query.
Together, $(\phi,g)$ form the search strategy, and we write $(\phi_t,g_t)=\Pi(\mathcal{D}_t)$ for how the strategy is built from the history at query $t$.

\begin{wrapfigure}{r}{0.44\textwidth}
\centering
\includegraphics[width=\linewidth]{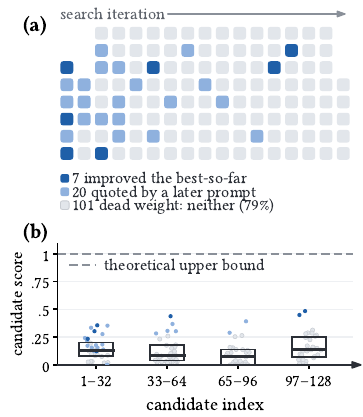}
\vspace{-1.5em}
\caption{\textbf{MCTS-style reward search:}
\textbf{(a)}~Selection coverage \textbf{(b)}~Candidate score.}
\label{fig:fixed-rule}
\vspace{-1em}
\end{wrapfigure}
\textbf{Limitations of fixed search rules.}
Predefined selection and proposal rules can restrict how a search uses accumulated feedback.
Existing LLM reward search fixes $\Pi$ by hand before the search, so every query uses the same pair.
Here, $\phi$ \emph{selects} which past evaluations condition the next query, by best-candidate selection, rank-based sampling, or Upper Confidence bounds applied to Trees (UCT)~\citep{uct}, and $g$ \emph{expands} the selection with edits drawn from a preset operator menu at fixed rates:
\begin{equation}
  R_{t+1}\sim \underbrace{g_{\mathcal{A}}\big(\cdot\mid\mathcal{T},\,\phi(\mathcal{D}_t)\big)}_{\text{one LLM generation}},
  \label{eq:baseline}
\end{equation}
where the selection criteria and proposal templates are predefined, as detailed in \Cref{tab:app-selection-expansion}. The templates may still condition generation on candidate code and feedback.
\Cref{fig:fixed-rule} measures both shortfalls on a logged MCTS-style run on DoorCloseOutward with $128$ samples, configured as in Appendix~\ref{sec:app-baselines}. \Cref{fig:fixed-rule}a marks each evaluated candidate by whether any later proposal reused it.
The score-based rule carries forward only a handful, and the code, the per-term reward curves, and the tracebacks of the other 79\% enter no later prompt in this run. 
Pooled over seven Bi-DexHands tasks, $77\%$ of the $684$ candidates neither improve the incumbent nor enter a later prompt.
Failed and low-scoring candidates are therefore unavailable for explicit reuse when the next candidate is written. \Cref{fig:fixed-rule}b shows limited improvement in candidate quality over the budget, with search performance plateauing below the maximum task score in this run. 
Learning $\Pi$ instead~\citep{metabox} would require meta-training over many problem instances, where each is evaluated by full RL training runs.

\vspace{-0.45\baselineskip}
\section{System Design}
\label{sec:method}
\vspace{-0.45\baselineskip}
This section presents the system design of \sys{}. We first specify the history representation and the proposal operator, with which the agent reads the history and writes reward candidates. Second, we specify the oracle and the exploration schedule, which remain outside the agent. We then assemble the four components into the complete optimization loop.

\begin{figure}[t]
\centering
\includegraphics[width=\textwidth]{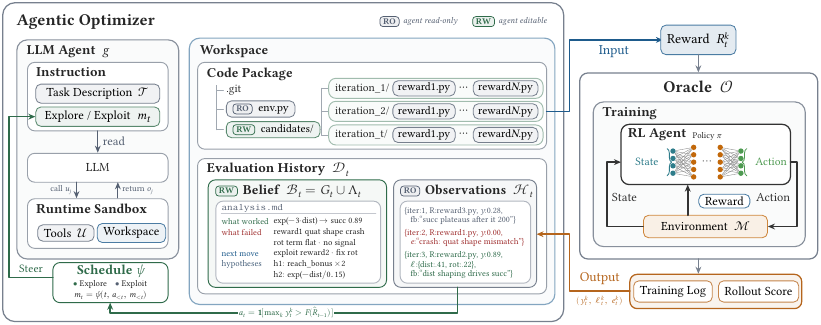}
\caption{\textbf{\sys{} framework design.} The agent queries the history $\mathcal{D}_t$ with tools and commits reward candidates, while the oracle owns the observations and the schedule issues the prompt $m_t$.}
\label{fig:architecture}
\end{figure}

\textbf{System overview.}
\sys{} performs the black-box optimization of \Cref{eq:generic} with four components, the history $\mathcal{D}_t$, the proposal operator $g$, the evaluation oracle $\mathcal{O}$, and the schedule $\psi$, as \Cref{fig:architecture} shows.
In each round, \sys{} performs scheduling, proposal, and evaluation in sequence:
\begin{equation}
  m_t=\psi\big(t,a_{<t},m_{<t}\big),\qquad
  \{R_t^k\}_{k=1}^{N_t}\sim g\big(\cdot\mid\mathcal{T},\mathcal{D}_t,m_t\big),\qquad
  \big(y_t^k,\ell_t^k,e_t^k\big)=\mathcal{O}\big(R_t^k\big),
  \label{eq:loop}
\end{equation}
where $t$ now indexes rounds, $m_t$ is the steering prompt of round $t$, $a_{<t}=(a_1,\dots,a_{t-1})$ the acceptance record, $m_{<t}$ the past steering prompts, $\{R_t^k\}_{k=1}^{N_t}$ the batch of candidates the agent commits, and $(y_t^k,\ell_t^k,e_t^k)$ the score, the per-term curves and the traceback of candidate $R_t^k$.
The schedule emits the steering prompt from the acceptance record, the agent generates the batch conditioned on the task specification and the history, and the oracle trains and scores every candidate, appending the results to the history.
Thus, the agent builds $(\phi_t,g_t)$ of \Cref{sec:formulation} at every round, selecting what it reads and expanding it into the next candidates.

\textbf{History representation.}
We remove compression from the stored history, and $\mathcal{D}_t$ retains the evaluation history in full, in two parts:
\begin{equation}
  \mathcal{D}_t=\underbrace{\mathcal{H}_t}_{\text{observations}}\ \cup\ \underbrace{\mathcal{B}_t}_{\text{belief}},
  \qquad
  \mathcal{B}_t=G_t\ \cup\ \Lambda_t,
  \label{eq:memory}
\end{equation}
where $\mathcal{H}_t$ are the observations maintained outside, $\mathcal{B}_t$ the belief updated by the agent itself, $G_t$ the agent's rolling notes, and $\Lambda_t$ the lineage of its edits.

The observations $\mathcal{H}_t$ are the oracle's side of the history, the evaluation history of \Cref{eq:generic} kept whole and grown only by union:
\begin{equation}
  \mathcal{H}_t=\{(R_r^k,y_r^k,\ell_r^k,e_r^k)\}_{r<t},
  \qquad
  \mathcal{H}_{t+1}=\mathcal{H}_t\ \cup\ h_t,
  \qquad
  h_t=\big\{\big(R_t^k,y_t^k,\ell_t^k,e_t^k\big)\big\}_{k=1}^{N_t},
  \label{eq:observations}
\end{equation}
where $r<t$ runs over the rounds before $t$ and $k$ over their candidates, each entry pairs $R_r^k$ with the score $y_r^k$, the per-term curves $\ell_r^k$ and the traceback $e_r^k$, and $h_t$ is the batch of records the round appends.
The oracle alone maintains them, and no agent action can revise them.

The belief $\mathcal{B}_t$ is the agent's side, and only the agent writes it:
\begin{equation}
  G_{t+1}\sim\mathrm{LLM}_\theta\big(\cdot\mid G_t,\ \mathcal{H}_t\big),
  \qquad
  \Lambda_{t+1}=\Lambda_t\ \cup\ \big\{(R_t^k,\,p_t)\big\}_{k=1}^{N_t}.
  \label{eq:belief}
\end{equation}
Each round the agent overwrites $G_t$ with its diagnoses and its intent, kept short since the full observations stay in $\mathcal{H}_t$.
The belief records hypotheses and search decisions that guide subsequent proposals, serving as an analogue of a posterior in Bayesian optimization.
The candidates are committed into $\Lambda_t$ with their parent commit $p_t$, on the best line when refining and on a fresh branch when exploring, a placement that records the agent's judgment of which line is worth extending and hence belongs to the belief. 

\textbf{Proposal operator.}
We implement $g$ as one full run of an autonomous agent, the frozen model $\mathrm{LLM}_\theta$ driving a sandboxed tool set $\mathcal{U}$ of bash, file operations and search~\citep{react,toolformer}, under a fixed instruction file, as Appendix~\ref{sec:app-skill} details.
The run works in the workspace of \Cref{fig:architecture}, where the environment code of $\mathcal{T}$ sits read-only in the code package, the candidates of each round are committed into it under git, the observations are the appended evaluation log, and the belief is the analysis file.
The run conditions on these files through its tool calls:
\begin{equation}
  u_{j+1}\sim\mathrm{LLM}_\theta\big(\cdot\mid c_t,\,\tau_j\big),
  \qquad
  o_{j+1}=\mathcal{U}\big(u_{j+1},\,\mathcal{T},\,\mathcal{D}_t\big),
  \qquad
  \tau_j=(u_1,o_1,\dots,u_j,o_j),
  \label{eq:operator}
\end{equation}
where $c_t$ is the fixed context of round $t$, holding the instruction, the task description and the steering prompt $m_t$ of \Cref{eq:schedule}, $u_j$ a tool call, and $o_j$ its observation of the workspace files.
A typical run reads the recorded scores, inspects policy training curves for the incumbent reward, checks the traceback of a failed candidate, and writes the next batch of candidates. 

The run ends with a tool call that commits the batch of candidates into the code package, and chaining the steps of \Cref{eq:operator} gives the proposal operator of \Cref{eq:loop}:
\begin{equation}
  g\big(\{R_t^k\}_{k=1}^{N_t},\,\tau\mid\mathcal{T},\mathcal{D}_t,m_t\big)
  =\prod_{j\ge 0}\mathrm{LLM}_\theta\big(u_{j+1}\mid c_t,\,\tau_j\big),
  \label{eq:agent-mdp}
\end{equation}
where the product runs over the tool calls of $\tau$, and the last of them commits the batch $\{R_t^k\}_{k=1}^{N_t}$.
The batch results from tool interactions with the workspace, rather than a single generation call.
Between rounds, the agent retains state through the history $\mathcal{D}_t$, which informs subsequent proposals.

\textbf{Oracle.}
In black-box optimization, each candidate is evaluated by an oracle, and a classical zeroth-order oracle returns only the scalar objective value~\citep{zerothorder}. Our oracle returns richer feedback:
\begin{equation}
  \mathcal{O}(R)=\big(F\big(A_{\mathcal{M}}(R)\big),\ \ell(R),\ e(R)\big),
  \qquad
  \mathcal{O}\notin\mathcal{U},
  \label{eq:oracle}
\end{equation}
where $A_{\mathcal{M}}$ trains one policy per committed candidate under the domain's own published configuration, $F$ is the task's acceptance metric, $\ell$ the per-term reward curves over training, $e$ the traceback when a candidate fails to run, and the triple is appended to $\mathcal{H}_t$.
The oracle runs outside the agent's environment, and no tool in $\mathcal{U}$ can invoke it.

\textbf{Exploration-Exploitation Schedule.}
The exploration-exploitation schedule is controlled externally. An unconstrained agent may overexploit the incumbent, and its belief provides no calibrated uncertainty estimates to guide exploration. 
The schedule $\psi$ therefore imposes the trade-off as the only scripted decision rule in the system, and it reads only the acceptance record and its own past prompts, carrying no knowledge of what the candidates contain:
\begin{equation}
  a_t=\mathbf{1}\Big[\max\nolimits_k y_t^k>F\big(\hat R_{t-1}\big)\Big],
  \qquad
  m_t=\psi\big(t,a_{<t},m_{<t}\big),
  \label{eq:schedule}
\end{equation}
where $a_t=1$ indicates that the round produces a candidate with a higher score than the incumbent reward $\hat R_{t-1}$.
The steering prompt takes one of two fixed values, $m_{\mathrm{exploit}}$ and $m_{\mathrm{explore}}$, each injected into the agent's context, directing the agent to extend $\hat R_{t-1}$ or to open a fresh line from scratch, and Appendix~\ref{sec:app-skill} shows the injected text.
Specifically, the schedule opens with $n_0$ rounds of $m_{\mathrm{explore}}$, holds $m_{\mathrm{exploit}}$ once an incumbent exists, and returns to $m_{\mathrm{explore}}$ for at most $\nu$ rounds after $a_t=0$ holds for $\mu$ consecutive rounds.

\textbf{Optimization loop.}
The search starts from the bootstrap code base with an empty history and repeats the round of \Cref{eq:loop} under the schedule, until the sample budget is spent. The search returns the incumbent reward, the highest-scoring candidate in the history.
\Cref{alg:search} in Appendix~\ref{sec:app-algorithms} shows the full reward discovery procedure of \sys{}.

\vspace{-0.45\baselineskip}
\section{Experiments}
\label{sec:experiments}
\vspace{-0.45\baselineskip}
This section describes the four control domains and the five automated baselines, and reports the comparison results. It then presents ablation studies on each component and backbone models.

\subsection{Experiment Environments}\vspace{-0.45\baselineskip}
\label{sec:exp-env}
We test \sys{} in four control domains, each with its own acceptance metric. Appendix~\ref{sec:app-environments} gives the score definitions and the sources of the task descriptions.\\ 
\textbf{Dexterous manipulation.} The ten dual-arm Bi-DexHands tasks~\citep{bidexhands} in Isaac Gym~\citep{isaacgym} cover handover and catching, articulated-object manipulation, and pick-and-place, stacking and pouring. Success is binary, and the score is the success rate. Appendix~\ref{sec:app-environments} lists all ten tasks.\\ 
\textbf{Power-grid operation.} The Case14 sandbox of Grid2Op~\citep{grid2op} is scored by the L2RPN scorer over $16$ held-out scenarios. Leading entries wrap RL in expert heuristics and simulator lookahead~\citep{l2rpn,l2rpnretro}, whereas we keep the task pure RL and place the whole operating objective on the reward.\\ 
\textbf{Locomotion.} The four Isaac Lab~\citep{isaaclab} tasks ask the ANYmal-D quadruped and the Cassie biped to track a commanded planar velocity and yaw rate on flat and on rough terrain, and the score charges tracking error and falls, where $0$ is perfect tracking.\\ 
\textbf{Traffic-signal control.} The RESCO benchmark~\citep{resco} runs SUMO~\citep{sumo} on real road networks with measured demand, one independent learner per intersection, over four networks. The score is the negated official average delay.

\vspace{-0.45\baselineskip}
\subsection{Baselines}
\label{sec:exp-baselines}
\vspace{-0.45\baselineskip}

We compare against five automated LLM search methods, each an instance of \Cref{eq:baseline}: a \emph{selection} rule fixed before the search begins decides which past candidates condition the next query, and a fixed \emph{expansion} rule $g_\mathcal{A}$ turns them into new candidates. The baseline methods include:

\textbf{Eureka~\citep{eureka}} selects the single best candidate of the previous batch and expands it with one improvement template. \\
\textbf{R$\star$~\citep{rstar}} reflects on the global best under a hint from \{improve, remove, add\} or crosses two rank-sampled parents, alternating with a parameter-alignment stage. \\
\textbf{REvolve~\citep{revolve}} maintains an evolving population of reward functions, selects parents from it by fitness, and expands them with mutation and crossover operations. We use its fully automated variant, where feedback comes from the environment rather than a human. \\
\textbf{ShinkaEvolve~\citep{shinka}} selects a score-weighted parent plus one archive and one top-$k$ inspiration and expands with a patch type drawn at fixed rates, testing whether a strong general program-evolution searcher closes the gap. \\
\textbf{RF-Agent~\citep{rfagent}} selects a node of its search tree by UCT, of which at most four enter any prompt, and expands it with a fixed menu of five actions.

\textbf{Settings.} We compare all methods at $72$ reward samples on locomotion and $128$ on the other three domains. Within each domain, methods share the evaluation budget, the \texttt{MiniMax-M2.7} backbone, and the policy-training configuration.
We report \emph{best-so-far} score against \emph{reward samples}, and \sys{} is labeled \emph{Ours} in every table and figure.
Appendix~\ref{sec:app-training} gives the full training settings.

\vspace{-0.45\baselineskip}
\subsection{Experiment Results}
\label{sec:exp-results}
\vspace{-0.45\baselineskip}

\textbf{Advantages of the agent under a shared budget.} \Cref{tab:results} reports each method's best-so-far success rate under the shared budget, where \emph{Sparse} uses the success indicator as the reward under the same policy training configuration as searched rewards. \sys{} attains the best score on most tasks.
It also never ranks below second among the automated methods on any of the ten tasks. 
Baseline performance varies across tasks. Individual baselines excel on some tasks but perform poorly on others, and none maintains an advantage across the suite. 
\sys{} maintains high success rates across tasks. 
Every method shares the same backbone and the same evaluation budget, and Appendix~\ref{sec:app-training} reports their LLM inference costs.

\begin{table}[H]
\centering
\caption{Bi-DexHands success rate within the $128$-sample budget, where \textbf{bold} and \underline{underline} mark the best and second-best automated search methods and runs short of the budget are reported at the samples completed.}
\label{tab:results}
\small
\setlength{\tabcolsep}{2.0pt}
\resizebox{\textwidth}{!}{%
\begin{tabular}{l*{10}{>{\centering\arraybackslash}p{3.4em}}|>{\centering\arraybackslash}p{3.0em}}
\toprule
Method & \multicolumn{4}{c}{\emph{Easy}} & \multicolumn{3}{c}{\emph{Medium}} & \multicolumn{3}{c}{\emph{Hard}} & \\
\cmidrule(lr){2-5}\cmidrule(lr){6-8}\cmidrule(lr){9-11}
& Swing \newline Cup & Lift \newline Underarm & Door \newline Close & Over & Grasp \newline Place & Block \newline Stack & Bottle \newline Cap & Kettle & Catch \newline Underarm & Catch \newline Abreast & \textbf{Avg} \\
\midrule
\emph{Sparse} & 0.01 & 0.00 & 0.06 & 0.00 & 0.00 & 0.02 & 0.03 & 0.02 & 0.00 & 0.00 & 0.014 \\
\midrule
REvolve & 0.920 & 0.906 & 0.606 & 0.947 & 0.510 & 0.323 & 0.695 & 0.970 & 0.298 & 0.372 & 0.655 \\
Eureka & \underline{0.994} & 0.889 & \underline{0.972} & 0.939 & \underline{0.527} & 0.529 & \underline{0.873} & 0.965 & 0.676 & 0.558 & \underline{0.792} \\
RF-Agent & 0.984 & \underline{0.930} & 0.484 & 0.925 & 0.366 & 0.170 & 0.837 & \underline{0.994} & 0.819 & \underline{0.560} & 0.707 \\
R$\star$ & \textbf{0.999} & 0.918 & 0.580 & \underline{0.970} & 0.339 & 0.352 & 0.434 & 0.958 & \textbf{0.850} & 0.000 & 0.640 \\
ShinkaEvolve & 0.879 & 0.738 & 0.775 & 0.878 & 0.330 & \textbf{0.917} & 0.609 & 0.986 & 0.753 & 0.541 & 0.741 \\
\midrule
Ours & \underline{0.994} & \textbf{0.979} & \textbf{1.000} & \textbf{0.971} & \textbf{0.984} & \underline{0.812} & \textbf{0.988} & \textbf{0.999} & \underline{0.849} & \textbf{0.603} & \textbf{0.918} \\
\bottomrule
\end{tabular}}
\vspace{-1em}
\end{table}

\begin{figure}[!htbp]
\centering
\includegraphics[width=\textwidth]{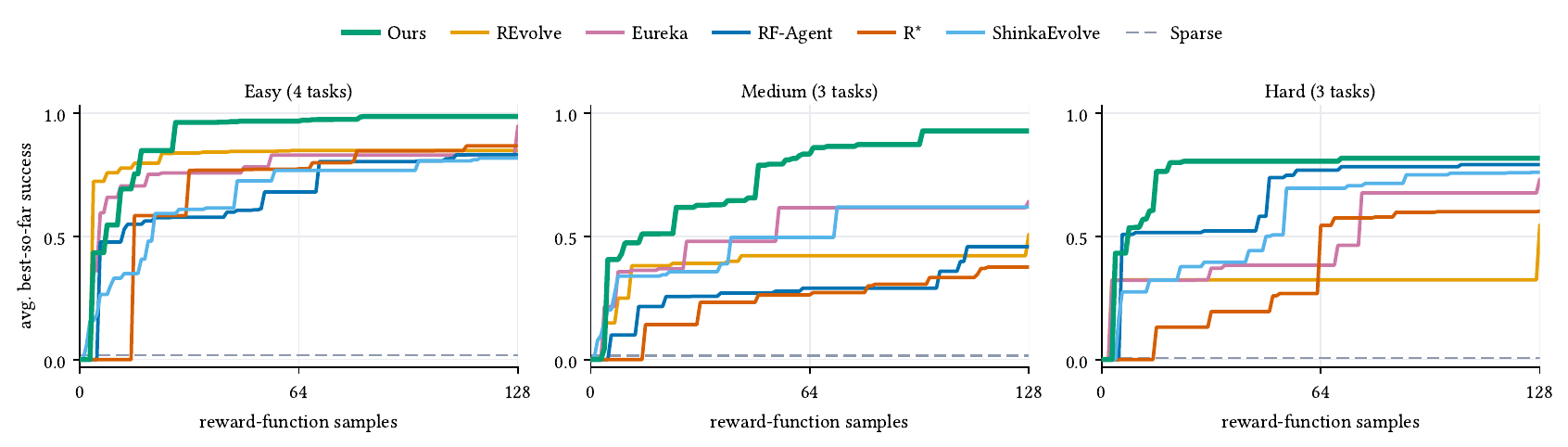}
\vspace{-1.5em}
\caption{Bi-DexHands sample efficiency by task difficulty, as group-mean best-so-far success against reward samples, with the group-mean \emph{Sparse} reference dashed.}
\label{fig:results}
\end{figure}

\textbf{Better rewards with fewer evaluations.} \Cref{fig:results} groups the ten tasks by difficulty and plots group-mean best-so-far success rates against reward samples. \sys{} achieves the highest final score in every group, and its score at $64$ samples already exceeds every baseline's final score. The gap is largest on the medium group, where it grows throughout the budget. On the hard group, \sys{} reaches its plateau within about $30$ samples, while the baselines close most of the gap by the end of the budget. On the easy group, every method ends above $0.8$ and the final gap is small.

\begin{figure}[!htbp]
\centering
\includegraphics[width=\textwidth]{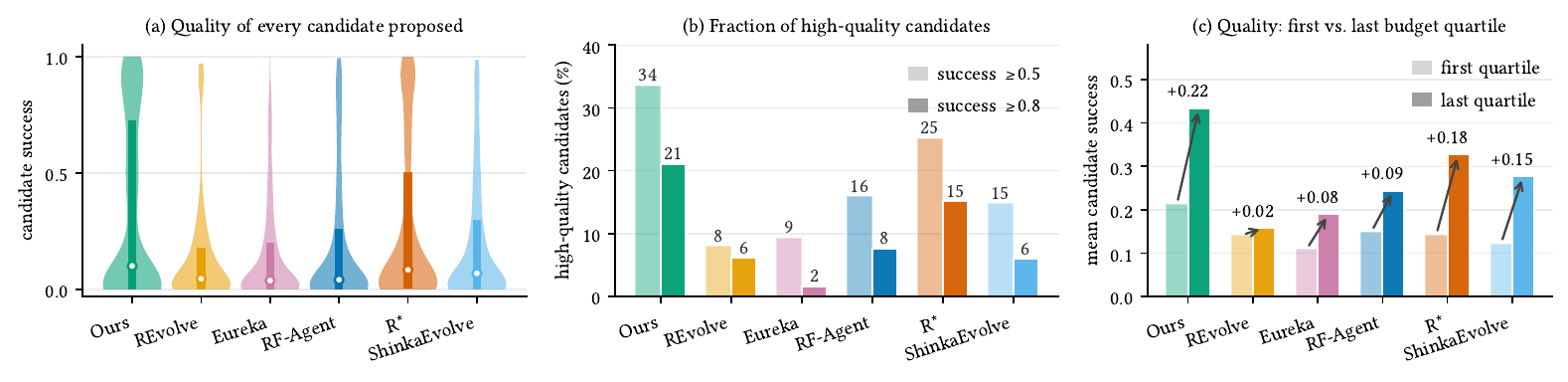}
\vspace{-1.5em}
\caption{Bi-DexHands candidate statistics per method: (a) success of all proposed candidates, (b) fractions clearing $0.5$ and $0.8$, (c) mean success in the first versus last budget quartile.}
\label{fig:candidates}
\end{figure}

\textbf{Autonomy generates better candidates.} \Cref{fig:candidates} summarizes the quality of all logged candidates. In \Cref{fig:candidates}a, the success rate distribution of \sys{}'s candidates dominates those of the fixed-rule baselines. In \Cref{fig:candidates}b, a larger fraction of its candidates exceeds each quality threshold than for any baseline. 
In \Cref{fig:candidates}c, its mean candidate quality increases most between the first and last budget quartiles, while even the strongest fixed-rule baseline improves less and the others still less.
Thus, \sys{} proposes higher-quality candidates on average, allocating fewer costly evaluations to low-quality candidates.
This improvement during search stems from the agent's autonomy.
The agent generates candidates after analyzing accumulated records, allowing subsequent batches to incorporate feedback from earlier failures. The predefined proposal operator $g_\mathcal{A}$ cannot match this improvement in candidate quality.

\begin{figure}[b]
\centering
\includegraphics[width=\textwidth]{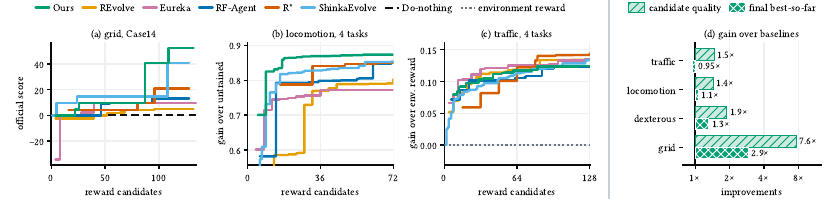}
\vspace{-1.5em}
\caption{Search progress on (a) Case14 by official score, and by mean relative gain over (b) the untrained policy on locomotion and (c) the environment reward on traffic. (d) \sys{}'s gain over the baseline mean in candidate quality and final best-so-far, domains from dense to sparse feedback.}
\label{fig:cps-methods}
\end{figure}

\textbf{The advantage widens on sparser-reward tasks.} \Cref{fig:cps-methods} tracks the search on the grid, locomotion, and traffic tasks. On traffic, with delay feedback at every step, all methods end at nearly the same level. On locomotion, \sys{} pulls ahead early and keeps the lead, though the baselines recover most of the gain. On the power grid, the sparsest setting, our reward scores highest and completes the most scenarios, and most baselines complete none. \Cref{fig:cps-methods}d adds manipulation and orders all domains from dense to sparse feedback. Candidate quality leads everywhere, and the final best-so-far dips to $0.95\times$ on traffic and reaches $2.9\times$ on the grid. The advantage is noticeable where feedback is sparse and unsuccessful candidates require diagnosis beyond their evaluation scores. Appendix~\ref{sec:app-case14-study} analyzes the grid states, and Appendices~\ref{sec:app-loco} and~\ref{sec:app-resco} report the locomotion and traffic runs.

\vspace{-0.45\baselineskip}
\subsection{Ablations}
\label{sec:ablation-mc}
\vspace{-0.45\baselineskip}

\begin{figure}[!htbp]
\centering
\vspace{-0.1em}
\includegraphics[width=\textwidth]{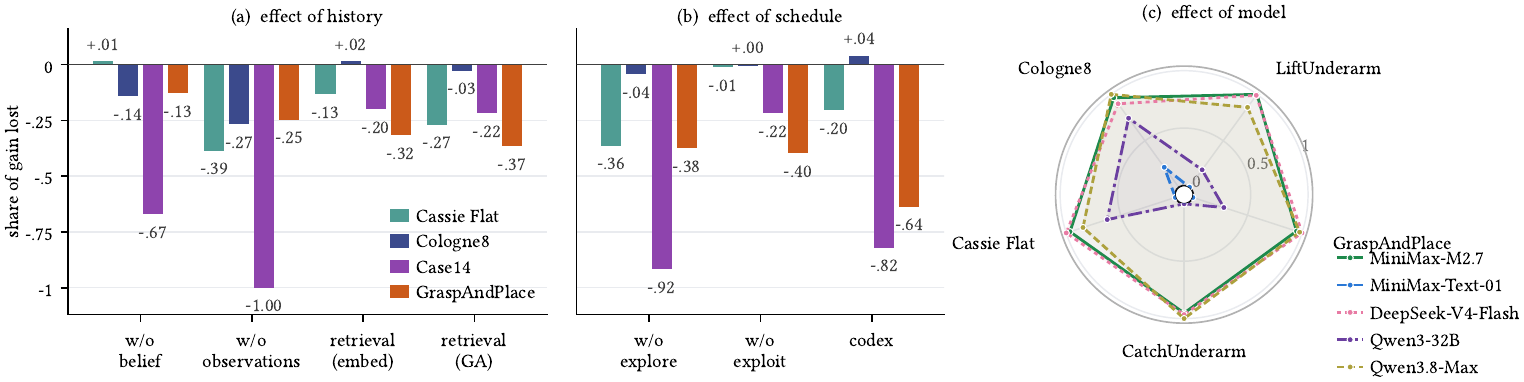}
\vspace{-0.9em}
\caption{Performance loss from ablating (a) the history, by removing the belief, the observations, or interposing a retriever, (b) the schedule, by using only exploration or exploitation or removing the framework, and (c) normalized score by backbone.}
\label{fig:ablation}
\end{figure}

\textbf{History Representation.} To evaluate access to historical information, we ablate each history component and the access policy. \emph{w/o belief} removes the self-maintained belief~\citep{reflexion}, and \emph{w/o observations} keeps only the bounded context of memory-managed agents~\citep{memgpt}. \emph{retrieval (embed)} and \emph{retrieval (GA)} select observations for the agent using embedding similarity~\citep{rag} or the retrieval rule of Generative Agents (GA)~\citep{generativeagents}. As shown in \Cref{fig:ablation}a, nearly every restriction on history access reduces search performance. Removing observations reduces performance more than removing the belief on every task, indicating their central role. Retrieval usually costs less than removing the observations but still limits performance gains. The search thus draws on evidence that score-based or similarity-based selection would omit.

\textbf{Exploration vs.\ Exploitation.} To evaluate the schedule $\psi$, we separately disable exploration and exploitation~\citep{suttonbarto} in the scheduling rule. \emph{w/o explore} always injects $m_{\mathrm{exploit}}$ and refines reward candidates from the best attempts, while \emph{w/o exploit} always injects $m_{\mathrm{explore}}$ and restarts every round. \emph{codex} removes the framework entirely and gives one stock coding agent the same task, backbone, and sample budget. As shown in \Cref{fig:ablation}b, the performance loss from using either strategy alone. On the easier traffic task every variant matches the full system, since dense feedback lets refinement and restarts converge on the same designs. On the harder power grid, \emph{w/o explore} and \emph{codex} give up most of the gain. This indicates that the schedule is dispensable where feedback is dense and essential where it is sparse.

\textbf{LLM Backbone.} To evaluate backbone effects, we run \sys{} with four alternative models. \Cref{fig:ablation}c reports the results. Scores are normalized using reference-reward performance and the highest backbone score for each task as shown in Appendix~\ref{sec:app-ablation-curves}. The non-reasoning backbone yields almost no performance gain. On most tasks, it exhausts the budget without a reward that supports successful policy training or outperforms the reference reward. Therefore, the proposed framework is not suited to weak backbones.

\vspace{-0.45\baselineskip}
\section{Limitation \& Conclusion}
\label{sec:conclusion}
\vspace{-0.45\baselineskip}
\textbf{Limitation.} \sys{} reasons over a purely textual history of scores, per-term curves, and tracebacks. Some failures are apparent only from non-textual observations, such as a policy achieving a high score through unnatural movements~\citep{rlvlmf} or a data center judged by its thermal fields~\citep{physicalai}. Extending the framework with multimodal reasoning is left to future work.

\textbf{Conclusion.} This paper proposes \sys{}, an agentic black-box optimization framework for automated reward discovery, in which an LLM agent builds the search strategy at run time from the evaluation history. The history is kept in two parts, the observations the oracle appends with the evaluation results and the belief the agent maintains with its diagnoses and intent. Experiments across diverse control tasks show performance gains over the evaluated baselines, and ablations support the contributions of the proposed components.

\clearpage
\subsection*{AI use statement}
We used generative AI tools to aid or polish writing, and for retrieval and discovery, which covers sourcing information and identifying relevant literature when surveying related work. We have reviewed all AI-assisted work. We take responsibility for the final content of this work, including text, claims or artifacts produced with the aid of generative AI.

\subsection*{Reproducibility Statement}
\Cref{sec:method} and Appendix~\ref{sec:app-algorithms} describe \sys{} and its full search procedure, and Appendix~\ref{sec:app-skill} gives all prompts. Appendices~\ref{sec:app-environments}--\ref{sec:app-baselines} specify the tasks, training configurations, seeds, sample budgets, schedule constants, and baseline settings.
Our code is available at \url{\coderepo}.

\subsection*{Ethics Statement}
This work involves no human subjects and no personal data, and all benchmarks are public simulators used under the licenses in Appendix~\ref{sec:app-licenses}. 

\bibliographystyle{iclr2027_conference}
\bibliography{references}

\clearpage
\appendix
\setcounter{topnumber}{4}\setcounter{bottomnumber}{3}\setcounter{totalnumber}{6}
\renewcommand{\topfraction}{0.95}\renewcommand{\bottomfraction}{0.95}
\renewcommand{\textfraction}{0.05}\renewcommand{\floatpagefraction}{0.85}
\section{Environments}
\label{sec:app-environments}
Every method reads the same natural-language description of each task. We do not reproduce these descriptions and instead cite the source that documents each benchmark.

\begin{figure}[!htbp]
\centering
\newcommand{\bdthumb}[2]{%
  \begin{minipage}[t]{0.192\textwidth}
    \centering
    \includegraphics[width=\linewidth]{figs/bidex_tasks/#1.jpg}\\[2pt]
    {\small #2}
  \end{minipage}}
\bdthumb{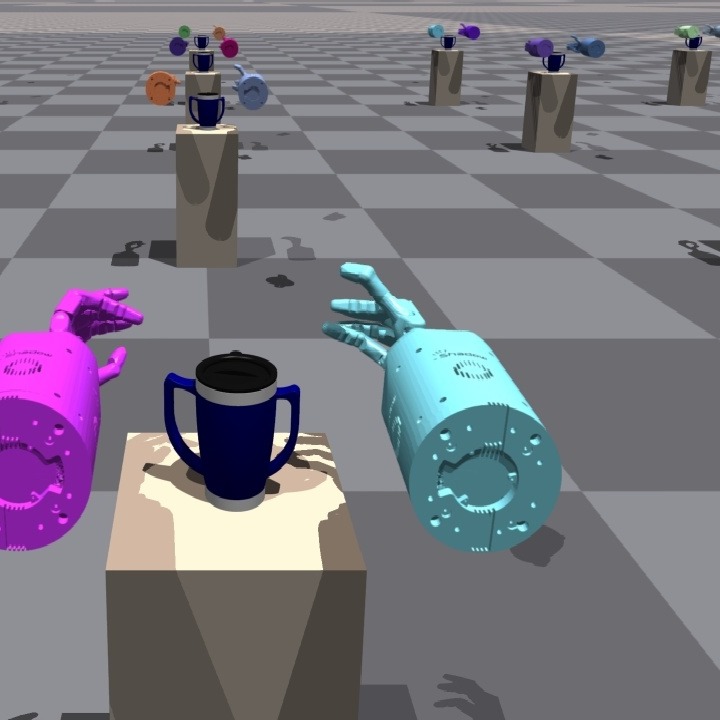}{SwingCup}\hfill
\bdthumb{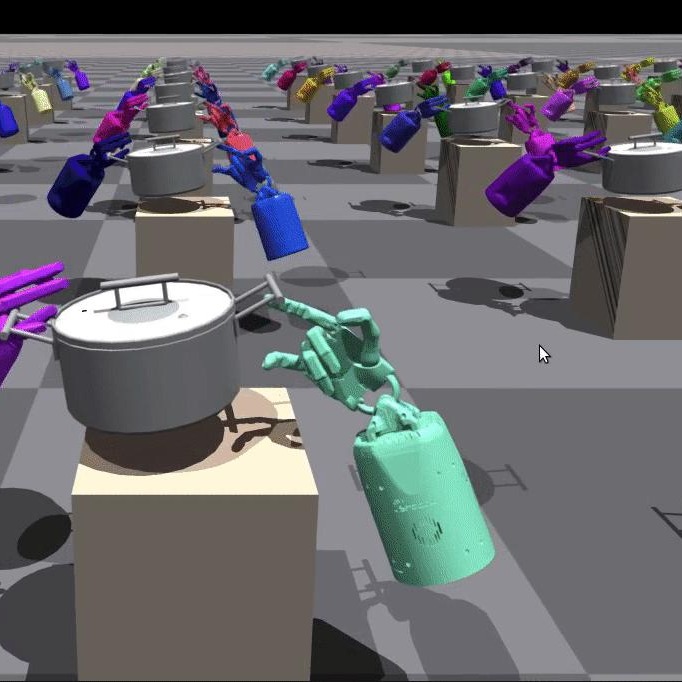}{LiftUnderarm}\hfill
\bdthumb{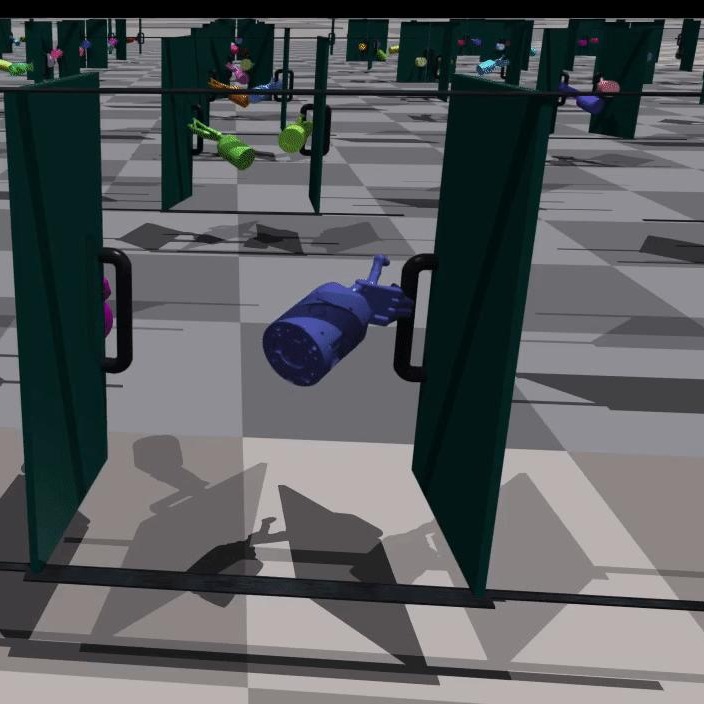}{DoorCloseOutward}\hfill
\bdthumb{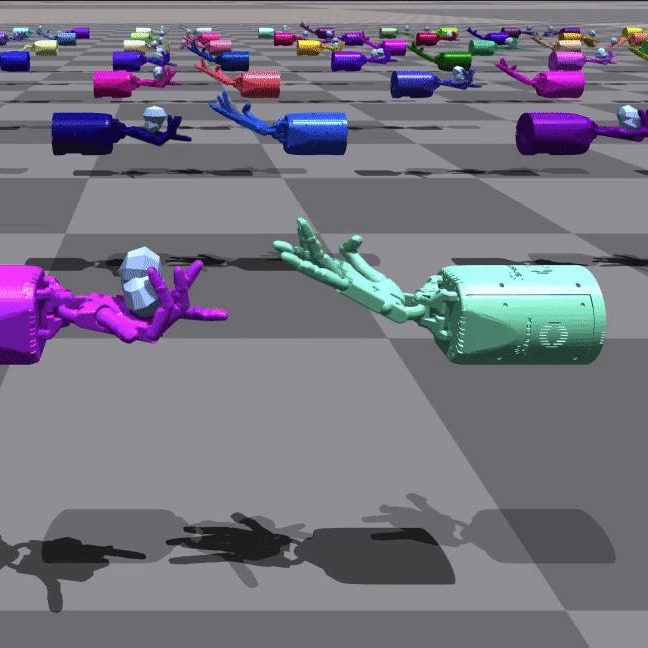}{Over}\hfill
\bdthumb{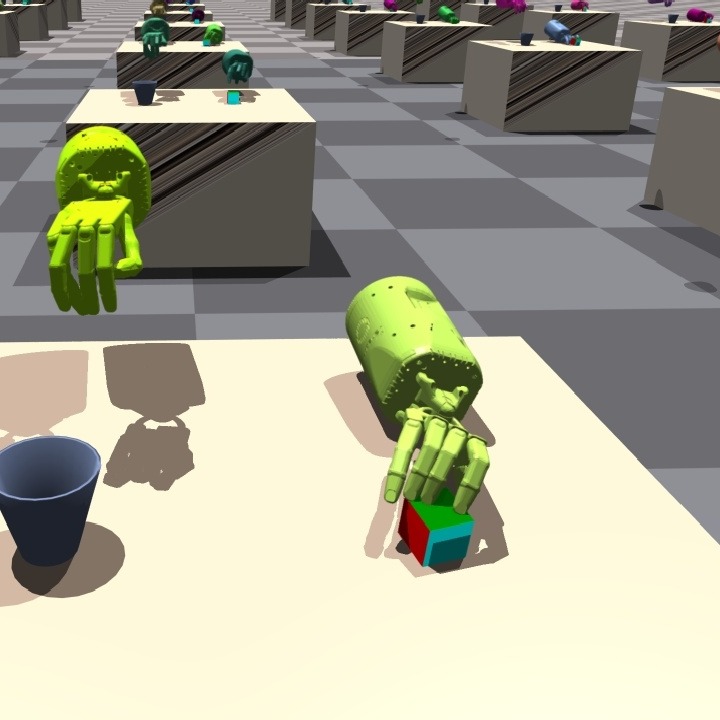}{GraspAndPlace}\\[0.2em]
\bdthumb{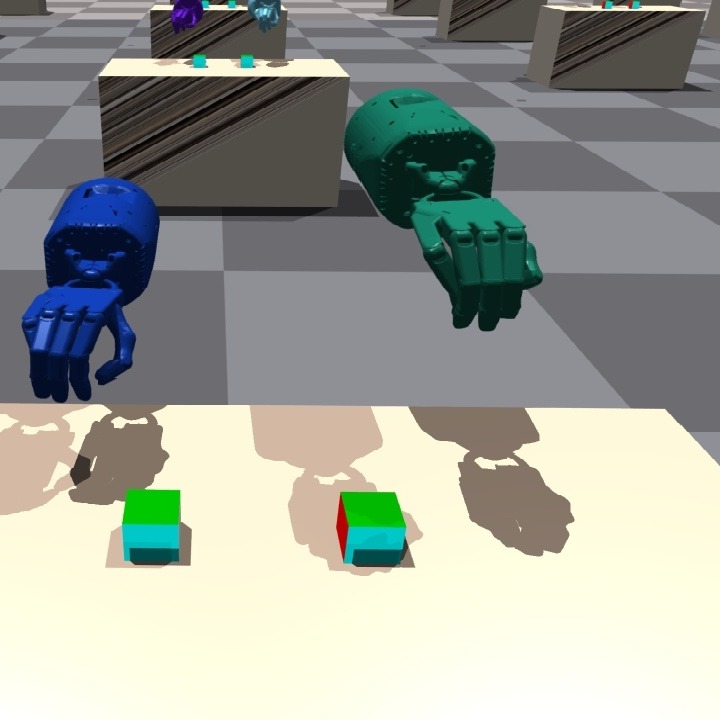}{BlockStack}\hfill
\bdthumb{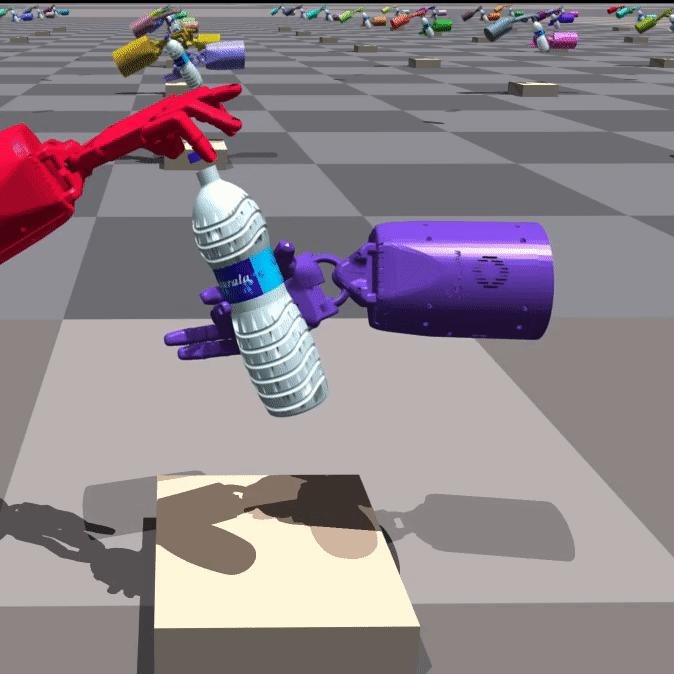}{BottleCap}\hfill
\bdthumb{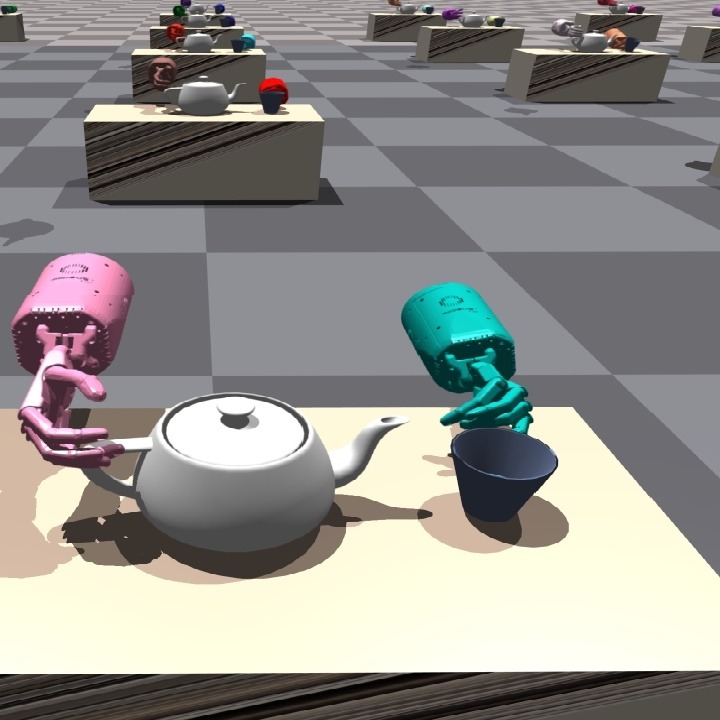}{Kettle}\hfill
\bdthumb{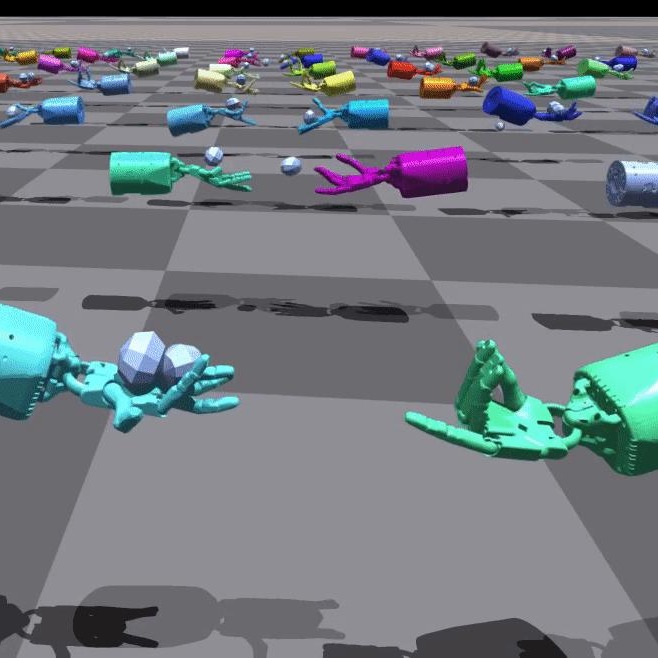}{CatchUnderarm}\hfill
\bdthumb{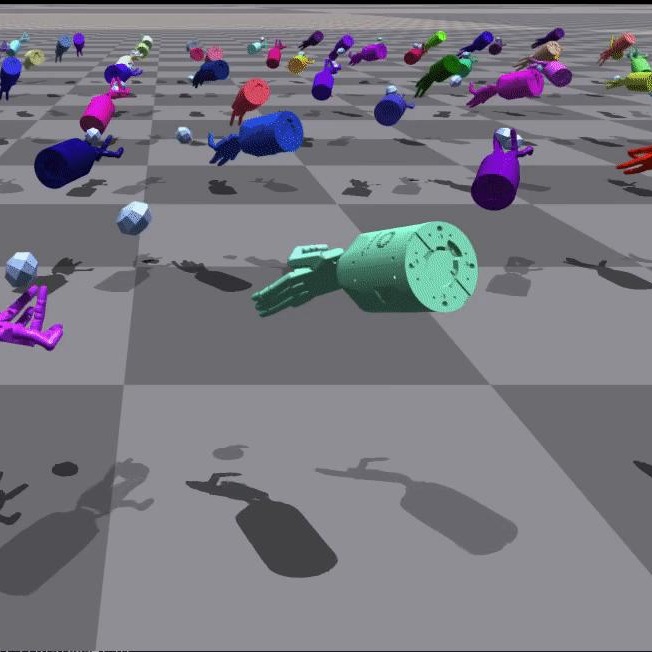}{CatchAbreast}
\caption{Bi-DexHands task environments.}
\label{fig:bidex-envs}
\end{figure}

\textbf{Bi-DexHands.} Bi-DexHands~\citep{bidexhands} places two shadow hands in Isaac Gym and asks them to cooperate, passing an object between hands, operating a tool, or placing a grasped object. Success is defined by a binary predicate over distances and orientations. The score therefore provides no partial credit. The dense reward supplies the learning signal. \Cref{fig:bidex-envs} illustrates the ten tasks, whose descriptions follow the original Bi-DexHands paper~\citep{bidexhands}.

\begin{figure}[!htbp]
\centering
\includegraphics[width=\textwidth]{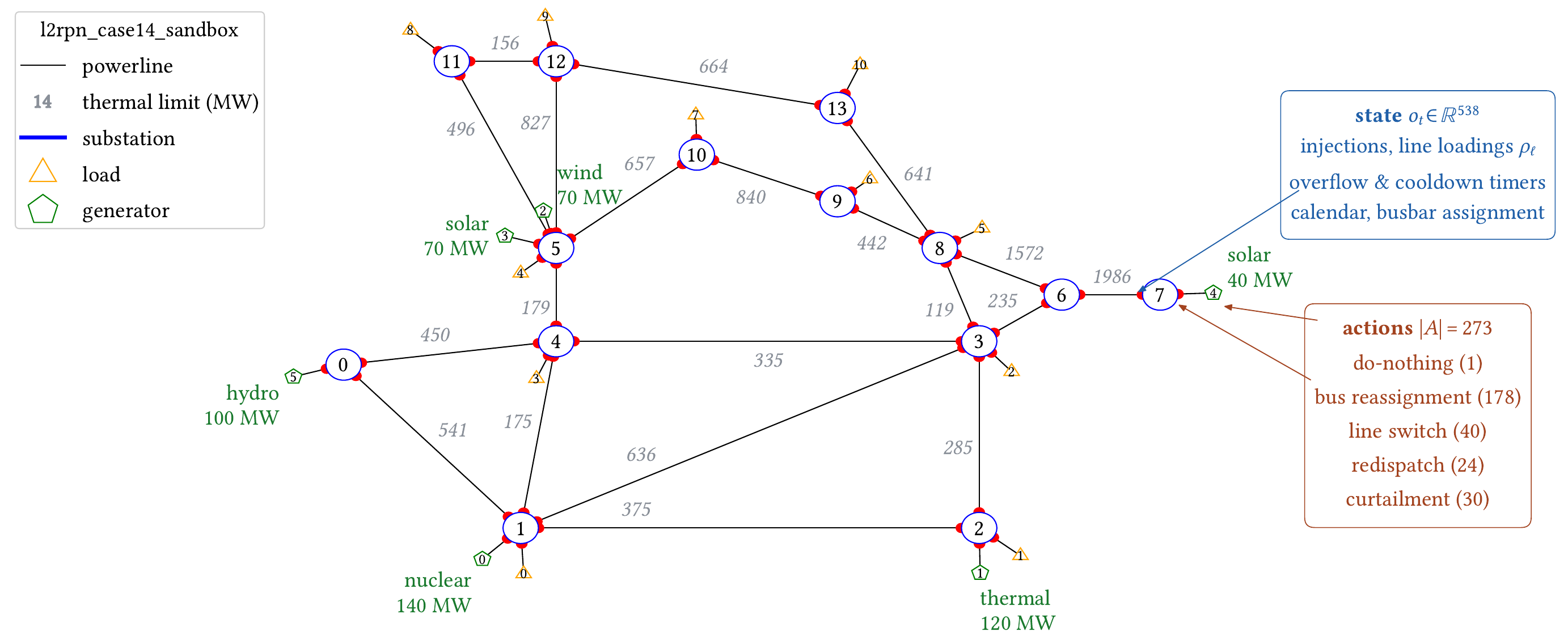}
\caption{L2RPN Case14 environment.}
\label{fig:grid-renders}
\end{figure}

\FloatBarrier
\textbf{Grid2Op.} Grid2Op~\citep{grid2op} simulates the real-time operation of a power transmission grid, held through four weeks of recorded demand by reconfiguring busbars, switching lines and redispatching generators. The score is the competition's operating cost rather than a success flag, pricing the grid survives and every step it does not. We use the L2RPN Case14 sandbox of \Cref{fig:grid-renders}, the IEEE case-14 benchmark~\citep{l2rpn}, which realizes the MDP $\langle S,A,T,R,\gamma,\rho_0\rangle$ as follows.
\vspace{-2pt}
\begin{itemize}\setlength{\itemsep}{0pt}
\item \textbf{State $S$.} The official Grid2Op state $o_t\in\mathbb{R}^{538}$: generator and load injections, per-line flows, loadings and status, overflow counters, cooldown timers, maintenance schedules, calendar features, and every element's busbar assignment.
\item \textbf{Actions $A$.} The $|A|=273$ competition controls: do-nothing ($1$), unitary single-substation bus reassignment ($178$), line disconnection and reconnection ($40$), unitary redispatch ($24$) and renewable curtailment ($30$).
\item \textbf{Transitions $T$.} Deterministic given the scenario: the simulator applies $a_t$, replays the next recorded injections, and re-solves the power flow. Writing $\rho_{\ell,t}=|f_{\ell,t}|/\bar f_\ell$ for the loading of line $\ell$ against its thermal limit, a line above $1$ for three consecutive steps or immediately above $2$ is disconnected and its flow redistributes onto neighbors, and the resulting cascade typically blacks out the grid at step $H_{\mathrm{alive}}\le H$.
\item \textbf{Initial state $\rho_0$ and horizon.} The initial state is the start of a scenario, which specifies a fixed sequence of injections over $H=8{,}064$ steps at five-minute resolution. Evaluation on the $16$ held-out scenarios use deterministic replays of these sequences.
\item \textbf{Reward $R$.} The object of search. Policies are trained with discount $\gamma=0.99$, and candidate selection uses the external evaluation score defined below.
\end{itemize}

Acceptance uses the L2RPN competition score~\citep{l2rpn}, which at every survived step charges the network losses plus the dispatch corrections actually applied, at the current marginal price,
\begin{equation}
c_t \;=\; p_t\,\Delta\Bigl(\;\underbrace{\sum\nolimits_{g} P^{\mathrm{gen}}_{g,t}
-\sum\nolimits_{d} P^{\mathrm{load}}_{d,t}}_{\text{network losses}}
\;+\; \underbrace{\sum\nolimits_{g} \bigl|P^{\mathrm{redisp}}_{g,t}\bigr|}_{\text{redispatch}}
\;+\; \underbrace{\bigl(P^{\mathrm{curt}}_{t}-P^{\mathrm{curt}}_{t-1}\bigr)}_{\text{curtailment change}}\;\Bigr),
\label{eq:grid-step-cost}
\end{equation}
Here $P^{\mathrm{gen}}_{g,t}$ and $P^{\mathrm{load}}_{d,t}$ are the produced and consumed active powers, $\Delta=\tfrac{1}{12}\,\mathrm{h}$ converts power to energy, $p_t$ is the cost of the most expensive generator producing, and $P^{\mathrm{redisp}}_{g,t}$ is the deviation applied to generator $g$ by the agent's requests and the environment's automatic balancing. The curtailment term prices the change in total curtailed renewable power $P^{\mathrm{curt}}_{t}$ and refunds released curtailment, and the official storage term is identically zero here. A blackout does not end the accounting: following the official implementation, every remaining step is charged the entire lost demand $D_t$, taken from the reference run and priced at $\bar p=\max_g p_g$,
\begin{equation}
C \;=\; \sum_{t=1}^{H_{\mathrm{alive}}} c_t
\;+\; \bar p \sum_{t=H_{\mathrm{alive}}+1}^{H} D_t .
\label{eq:grid-episode-cost}
\end{equation}
The episode score maps $C$ piecewise-linearly, clipping at both ends, through four anchor costs on the same scenario: an idealized loss-minimizing operation, the official no-overflow ``safe finish'' reference, Do-Nothing, and an immediate whole-scenario blackout with $C_{\mathrm{worst}}=\bar p\sum_{t=1}^{H} D_t$. The score $F$ averages it over the $16$ held-out scenarios,
\begin{equation}
S = \sigma(C), \quad
\sigma\colon \bigl(C_{\mathrm{best}},\,C_{\mathrm{reco}},\,C_{\mathrm{DN}},\,C_{\mathrm{worst}}\bigr)
\mapsto \bigl(100,\,80,\,0,\,-100\bigr),
\qquad
F = \frac{1}{16}\sum_{s=1}^{16} S_s .
\label{eq:grid-score}
\end{equation}
Completing a scenario more cheaply than the safe-finish reference therefore scores above $80$, and operating longer and cheaper than Do-Nothing scores above $0$. The task description below follows Grid2Op~\citep{grid2op} and the L2RPN competition~\citep{l2rpn}.

\smallskip\noindent{\small\begin{tabular}{@{}p{\linewidth}@{}}
\toprule
\emph{Case14} task description \\
\midrule
Keep every load supplied for the entire scenario. Avoid unnecessary switching, intervene only to relieve persistent overloads or restore connectivity, prioritizing blackout prevention before operating cost. Per-scenario operation cost, interpolated linearly between the best, safe-finish, do-nothing and worst references to $100$, $80$, $0$ and $-100$, then averaged over the $16$ held-out scenarios. \\
\bottomrule
\end{tabular}}
\smallskip

\begin{figure}[!htbp]
\centering
\newcommand{\rsthumb}[2]{%
  \begin{minipage}[t]{0.16\textwidth}
    \centering
    \includegraphics[width=\linewidth]{figs/resco_maps/#1.pdf}\\[2pt]
    {\small #2}
  \end{minipage}}
\rsthumb{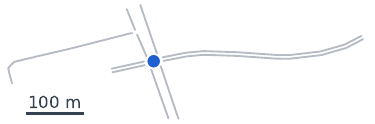}{Cologne1}\hspace{0.03\textwidth}
\rsthumb{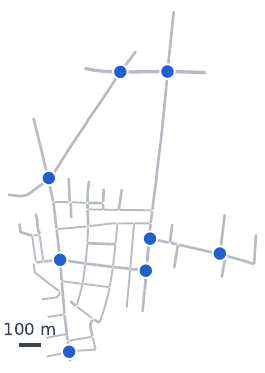}{Cologne8}\hspace{0.03\textwidth}
\rsthumb{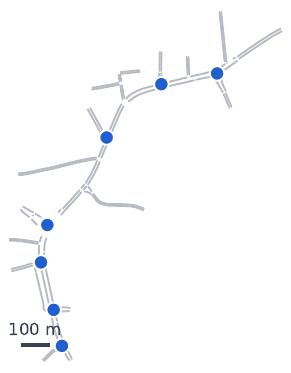}{Ingolstadt7}\hspace{0.03\textwidth}
\rsthumb{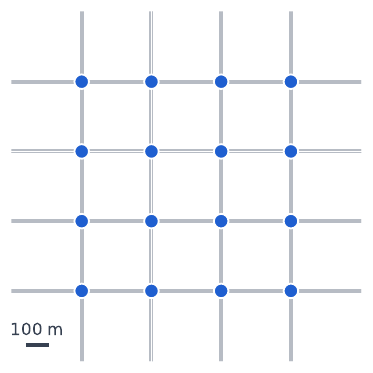}{Grid4x4}
\caption{RESCO traffic networks, with controlled signals in blue.}
\label{fig:resco-envs}
\end{figure}

\textbf{RESCO.} RESCO~\citep{resco} couples the SUMO traffic simulator~\citep{sumo} to real road networks and measured demand, and gives every signalized intersection its own learner that reads only its own entry of the designed reward. Because the learners do not communicate at execution time, the designed reward function must provide the incentives for coordination. The score $F=-\min_{e\le 40}\bar d_e$ is the negated average delay of the best training episode. The four networks of \Cref{fig:resco-envs} are a single real intersection, a real eight-signal corridor, an Ingolstadt district of $7$ controlled signals, and a synthetic $4{\times}4$ grid. Their (obs, action) dimensions are (40, 4), (165, 23), (290, 21) and (960, 128). RESCO provides no task descriptions, so we write one per network. Each states the network and its demand, the $10$\,s control step with a $3$\,s yellow before every phase change, that unserved vehicles stay queued in the network, and the goal of minimizing vehicle delay.

\begin{figure}[!htbp]
\centering
\newcommand{\lcthumb}[2]{%
  \begin{minipage}[t]{0.185\textwidth}%
    \centering
    \includegraphics[width=\linewidth]{figs/isaaclab_loco/#1.png}\\[1pt]
    {\fontsize{8.65}{10.4}\selectfont #2}%
  \end{minipage}}
\newcommand{\lcgap}{\hspace{0.012\textwidth}}
\lcthumb{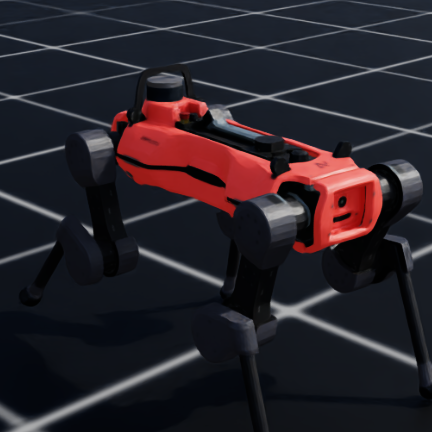}{flat}\lcgap%
\lcthumb{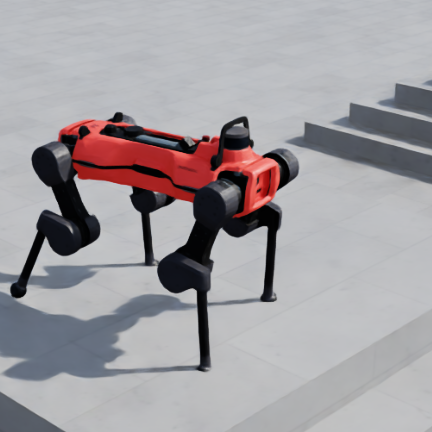}{stairs up}\lcgap%
\lcthumb{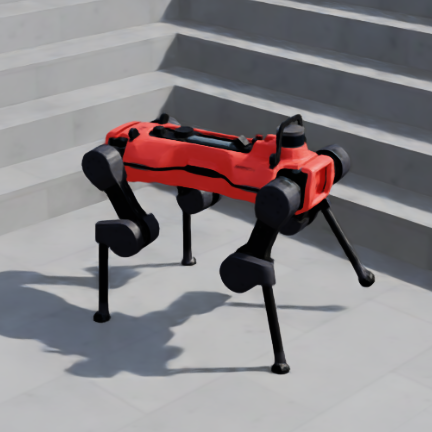}{stairs down}\lcgap%
\lcthumb{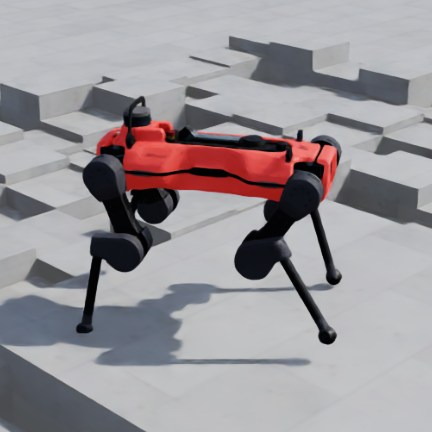}{boxes}\\[0.5em]
\lcthumb{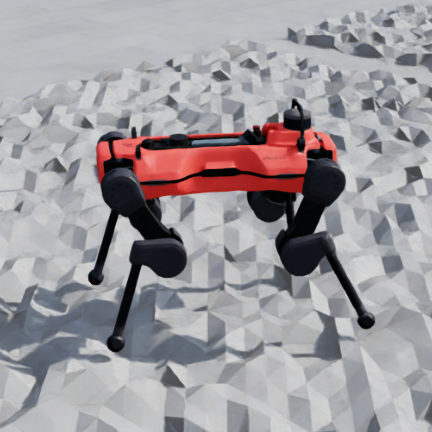}{random rough}\lcgap%
\lcthumb{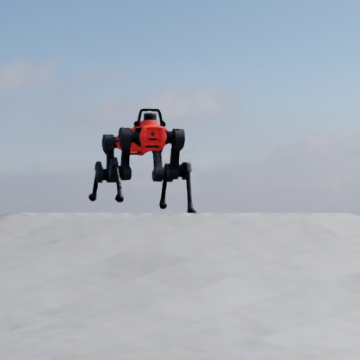}{slope up}\lcgap%
\lcthumb{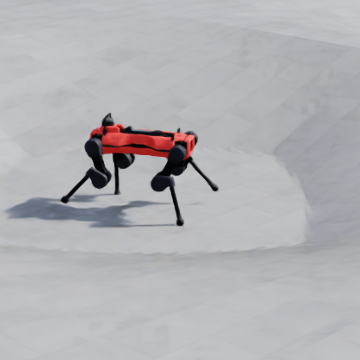}{slope down}\\[0.3em]
{\fontsize{8.65}{10.4}\selectfont ANYmal-D}\\[0.7em]
\lcthumb{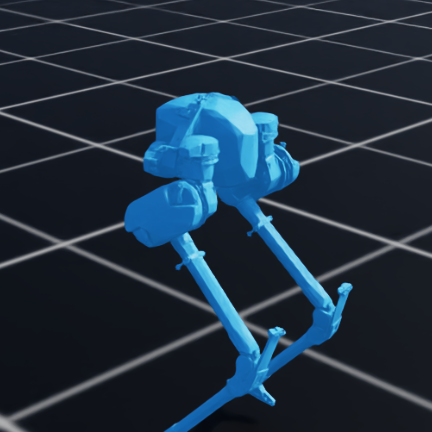}{flat}\lcgap%
\lcthumb{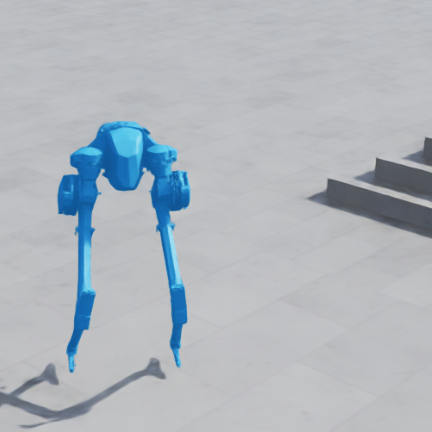}{stairs up}\lcgap%
\lcthumb{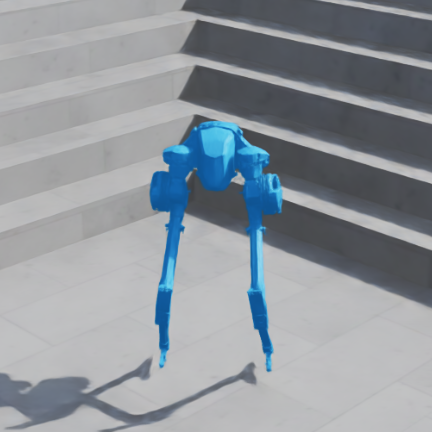}{stairs down}\lcgap%
\lcthumb{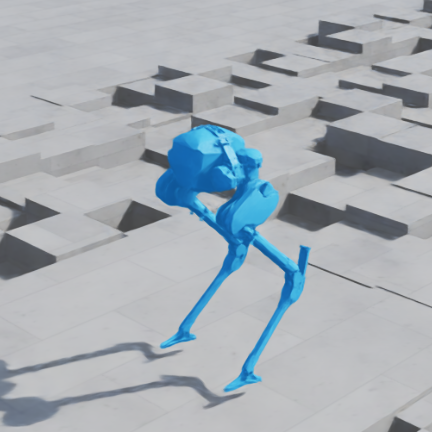}{boxes}\\[0.5em]
\lcthumb{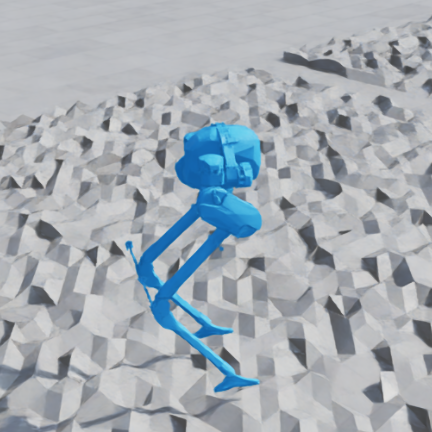}{random rough}\lcgap%
\lcthumb{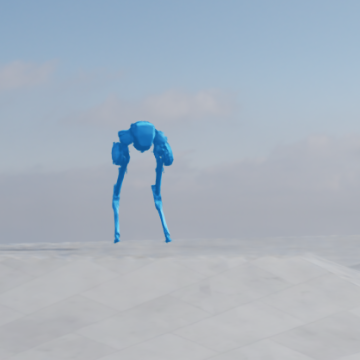}{slope up}\lcgap%
\lcthumb{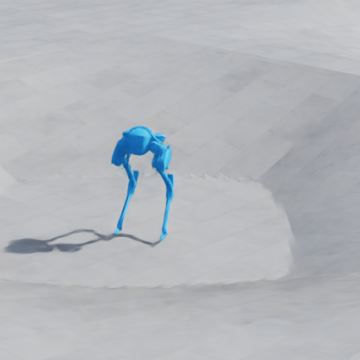}{slope down}\\[0.3em]
{\fontsize{8.65}{10.4}\selectfont Cassie}
\caption{Isaac Lab locomotion environments.}
\label{fig:loco-envs}
\end{figure}

\FloatBarrier
\textbf{Isaac Lab locomotion.} Isaac Lab~\citep{isaaclab} asks the ANYmal-D quadruped and the Cassie biped to hold a commanded velocity on flat and on rough terrain, cutting the episode short on any fall, and with no binary success the score is the tracking error plus a fall charge, which bills the commanded speed for every step a fall removes from the episode. The four tasks are illustrated in \Cref{fig:loco-envs}, and their descriptions follow Isaac Lab~\citep{isaaclab}.

\clearpage
\FloatBarrier
\section{Training Details}
\label{sec:app-training}
\paragraph{Policy training.} For each candidate reward, a policy is trained from scratch using the benchmark's published configuration. Only the reward function varies. \Cref{tab:app-ppo} lists the training configurations for the four oracles. All six methods search under training seed $0$, enforced on the server side. The selected best rewards are additionally trained with seeds $1$--$5$ for the curves in Appendix~\ref{sec:app-results}.

\begin{table}[h]
\centering
\caption{Policy-training configuration of the four oracles.}
\label{tab:app-ppo}
\small
\setlength{\tabcolsep}{3.5pt}
\begin{tabularx}{0.9\linewidth}{@{}>{\raggedright\arraybackslash\hsize=1.217\hsize}X
                  >{\raggedright\arraybackslash\hsize=0.467\hsize}X
                  >{\raggedright\arraybackslash\hsize=1.947\hsize}X
                  >{\raggedright\arraybackslash\hsize=0.759\hsize}X
                  >{\raggedright\arraybackslash\hsize=1.120\hsize}X
                  >{\raggedright\arraybackslash\hsize=0.857\hsize}X
                  >{\raggedright\arraybackslash\hsize=0.633\hsize}X@{}}
\toprule
& Algo. & Policy network & Learn. rate & Budget per candidate & Sim.\ step \\
\midrule
Bi-DexHands & PPO & MLP $512{-}256{-}128$ & $3\times10^{-4}$ & $3000$ iterations & $1/60$\,s \\
Grid2Op Case14 & PPO & MLP $256{-}256$ & $3\times10^{-4}$ & $10{,}240$ timesteps & $5$\,min \\
Isaac Lab locomotion & PPO & MLP $128{-}128{-}128$ (flat) to $512{-}256{-}128$ (rough) & $1\times10^{-3}$ & $800$ iterations & $0.005$\,s \\
RESCO & DQN & Conv $2{\times}2{\times}64$, MLP $64{-}64$ & $1\times10^{-3}$ & $40$ episodes & $10$\,s \\
\bottomrule
\end{tabularx}
\end{table}

\paragraph{Search configuration.} \Cref{tab:app-search} lists \sys{}'s search configuration for the main comparisons. The schedule holds $m_{\mathrm{explore}}$ for the first $n_0$ rounds, extended until an incumbent exists, exploits thereafter, and inserts a re-exploration burst of at most $\nu$ rounds once $\mu$ consecutive rounds go unaccepted. Each attempted candidate counts as one sample.

\begin{table}[H]
\centering
\caption{\sys{} search configuration for the main comparisons.}
\label{tab:app-search}
\small
\setlength{\tabcolsep}{6pt}
\begin{tabular}{@{}>{\raggedright\arraybackslash}p{0.24\textwidth} >{\raggedright\arraybackslash}p{0.1\textwidth}@{}}
\toprule
Parameter & Value \\
\midrule
Locomotion samples & $72$ \\
Other domains' samples & $128$ \\
Initial explore rounds, $n_0$ & $5$ \\
Exploit patience, $\mu$ & $20$ \\
Re-exploration burst, $\nu$ & $5$ \\
\bottomrule
\end{tabular}
\end{table}

\paragraph{Backbones.} \Cref{tab:app-backbones} lists the five backbones of \Cref{fig:ablation}c, chosen to vary the model family, the reasoning capability and the serving scale, each driven through the same OpenAI-compatible interface at the same temperature, instruction file, tools and schedule.

\begin{table}[H]
\centering
\caption{Backbone ablation configurations.}
\label{tab:app-backbones}
\small
\setlength{\tabcolsep}{5pt}
\begin{tabular}{@{}>{\raggedright\arraybackslash}p{0.2\textwidth} >{\raggedright\arraybackslash}p{0.26\textwidth} >{\raggedright\arraybackslash}p{0.13\textwidth}@{}}
\toprule
Backbone & Serving & Reasoning \\
\midrule
MiniMax-M2.7 & hosted API & yes \\
MiniMax-Text-01 & hosted API & no \\
DeepSeek-V4-Flash & hosted API & yes \\
Qwen3.8-Max & hosted API & yes \\
Qwen3-32B & local $4$-bit quantization & yes \\
\midrule
Temperature & \multicolumn{2}{l}{$0.7$ for every backbone} \\
\bottomrule
\end{tabular}
\end{table}

\paragraph{Experiment environment.} Agent, oracle and every training server share the single node of \Cref{tab:app-node}, where the Isaac Gym and Isaac Lab oracles train one batch in parallel on the GPUs while the Grid2Op and RESCO oracles train on the CPU threads.

\begin{table}[H]
\centering
\caption{Hardware and software configuration.}
\label{tab:app-node}
\footnotesize
\renewcommand{\arraystretch}{0.95}
\setlength{\tabcolsep}{6pt}
\begin{tabular}{@{}>{\raggedright\arraybackslash}p{0.16\textwidth} >{\raggedright\arraybackslash}p{0.56\textwidth}@{}}
\toprule
Environment  & Specification \\
\midrule
GPUs & $4\times$ NVIDIA GeForce RTX~4090, $48$\,GiB memory each \\
CPU & Intel Xeon Platinum 8173M, $112$ threads \\
Memory & $503$\,GiB \\
OS & Ubuntu 22.04.5 LTS \\
Software & NVIDIA driver $590.48.01$, CUDA $13.1$ \\
\bottomrule
\end{tabular}
\end{table}

\paragraph{Per-candidate training cost.} \Cref{tab:app-walltime} reports per-candidate training and evaluation times in minutes, with the mean and $10$th--$90$th percentile range for each task.

\begin{table}[H]
\centering
\caption{Per-candidate wall clock in minutes, mean and $10$th--$90$th percentile range.}
\label{tab:app-walltime}
\small
\setlength{\tabcolsep}{3pt}
\begin{tabular}[t]{@{}lcc@{}}
\toprule
Task & Mean & Range \\
\midrule
\multicolumn{3}{@{}l}{\emph{Bi-DexHands}} \\
SwingCup & $44$ & $11$--$62$ \\
LiftUnderarm & $73$ & $56$--$91$ \\
DoorCloseOutward & $65$ & $57$--$73$ \\
Over & $42$ & $38$--$47$ \\
GraspAndPlace & $23$ & $16$--$32$ \\
BlockStack & $60$ & $51$--$73$ \\
BottleCap & $70$ & $65$--$77$ \\
Kettle & $134$ & $111$--$162$ \\
CatchUnderarm & $46$ & $33$--$62$ \\
CatchAbreast & $42$ & $31$--$58$ \\
\bottomrule
\end{tabular}\hspace{1em}
\begin{tabular}[t]{@{}lcc@{}}
\toprule
Task & Mean & Range \\
\midrule
\multicolumn{3}{@{}l}{\emph{Grid2Op}} \\
Case14 & $11$ & $10$--$13$ \\
\midrule
\multicolumn{3}{@{}l}{\emph{Locomotion}} \\
ANYmal-D Flat & $72$ & $48$--$96$ \\
ANYmal-D Rough & $114$ & $87$--$139$ \\
Cassie Flat & $67$ & $37$--$106$ \\
Cassie Rough & $76$ & $57$--$97$ \\
\bottomrule
\end{tabular}\hspace{1em}
\begin{tabular}[t]{@{}lcc@{}}
\toprule
Task & Mean & Range \\
\midrule
\multicolumn{3}{@{}l}{\emph{RESCO}} \\
Cologne1 & $9$ & $7$--$10$ \\
Cologne8 & $22$ & $19$--$24$ \\
Ingolstadt7 & $28$ & $26$--$31$ \\
Grid4x4 & $61$ & $52$--$72$ \\
\bottomrule
\end{tabular}
\end{table}

\paragraph{LLM inference cost.} \Cref{tab:app-cost} reports each method's prefill and decode tokens per attempted candidate.

\begin{table}[H]
\centering
\caption{LLM tokens per attempted candidate, averaged over the ten Bi-DexHands tasks.}
\label{tab:app-cost}
\small
\setlength{\tabcolsep}{5pt}
\begin{tabular}{@{}lrr@{}}
\toprule
Method & Prefill Tokens & Decode Tokens \\
\midrule
\textbf{Ours} & \textbf{176{,}961} & \textbf{21{,}671} \\
RF-Agent & 10{,}783 & 9{,}990 \\
REvolve & 3{,}940 & 3{,}488 \\
Eureka & 3{,}919 & 3{,}511 \\
R$\star$ & 3{,}391 & 6{,}132 \\
ShinkaEvolve & 19{,}242 & 7{,}726 \\
\bottomrule
\end{tabular}
\end{table}

\FloatBarrier
\section{Baseline Details}
\label{sec:app-baselines}

Every baseline runs its own search from its released code, or from our reimplementation for R$\star$, against the oracle at identical settings, with reasoning traces stripped before each method's parser sees the reply. \Cref{tab:app-selection-expansion} summarizes each method's selection and expansion rule.

\begin{table}[!htb]
\centering
\caption{Selection and expansion rules of the compared methods.}
\label{tab:app-selection-expansion}
\footnotesize
\setlength{\tabcolsep}{4pt}
\renewcommand{\arraystretch}{1.05}
\begin{tabular}{@{}>{\raggedright\arraybackslash}p{0.19\textwidth} >{\raggedright\arraybackslash}p{0.38\textwidth} >{\raggedright\arraybackslash}p{0.35\textwidth}@{}}
\toprule
Method & Selection & Expansion  \\
\midrule
Eureka~\citep{eureka} & Previous batch's best, one per prompt. & One fixed improvement template. \\
REvolve~\citep{revolve} & Rank-weighted from an elite archive capped at $13$, one or two per prompt. & Mutation or crossover at equal rates. \\
RF-Agent~\citep{rfagent} & UCT over the tree, at most four nodes per prompt. & Fixed menu of five actions. \\
R$\star$~\citep{rstar} & Global best for reflection, two rank-sampled parents for crossover. & Reflection hint from \{improve, remove, add\} or module crossover, plus parameter alignment. \\
ShinkaEvolve~\citep{shinka} & Score-weighted parent, one archive and one top-$k$ inspiration, at most three. & Patch type from \{diff, full, cross\} at fixed rates. \\
\midrule
Ours & Chosen by the agent with tools over the full history, failures included. & Free-form edits from its diagnosis, batch size its own. \\
\bottomrule
\end{tabular}
\end{table}

\textbf{Eureka.} We set the per-iteration batch size to $4$.

\textbf{REvolve.} We use the fully automated variant without the human-feedback loop, and set $4$ candidates per iteration with the elite archive capped at $13$.

\textbf{RF-Agent.} We set the per-expansion action counts to $(2,2,2,1,1)$ over its five actions, eight children per expansion, and the UCT constant annealed linearly from $0.4$ to $0.1$, with thought-align and self-verify enabled.

\textbf{R$\star$.} In the absence of an official implementation, we reimplement the method from the paper. We use a population of $16$, generate $12$ reflection offspring and $4$ module-level crossover offspring per iteration, and employ $5$ LLM critics for Bradley-Terry parameter alignment. We retrain the top $2$ individuals at each iteration.

\textbf{ShinkaEvolve.} We set one candidate per generation, $2$ islands, an archive of $40$ with elite ratio $0.3$, weighted parent selection at $\lambda=10$, one archive and one top-$k$ inspiration per prompt, and diff, full and cross patch probabilities of $0.6$, $0.3$ and $0.1$, with the LLM novelty judge on, while the embedding-based novelty filter is disabled.

\section{Additional Results}
\label{sec:app-results}

\subsection{Dexterous Manipulation}
\label{sec:app-bidex}
\begin{figure}[!htbp]
\centering
\includegraphics[width=\textwidth]{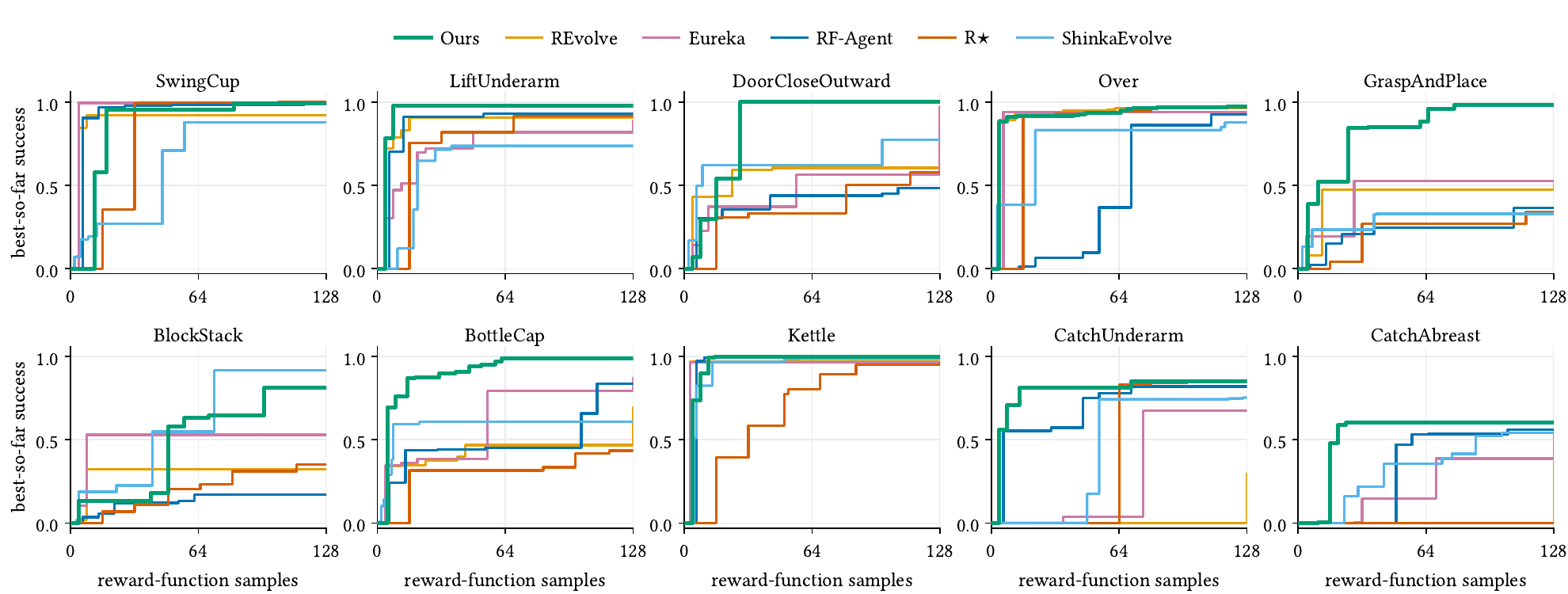}\\[0.1em]
\includegraphics[width=\textwidth]{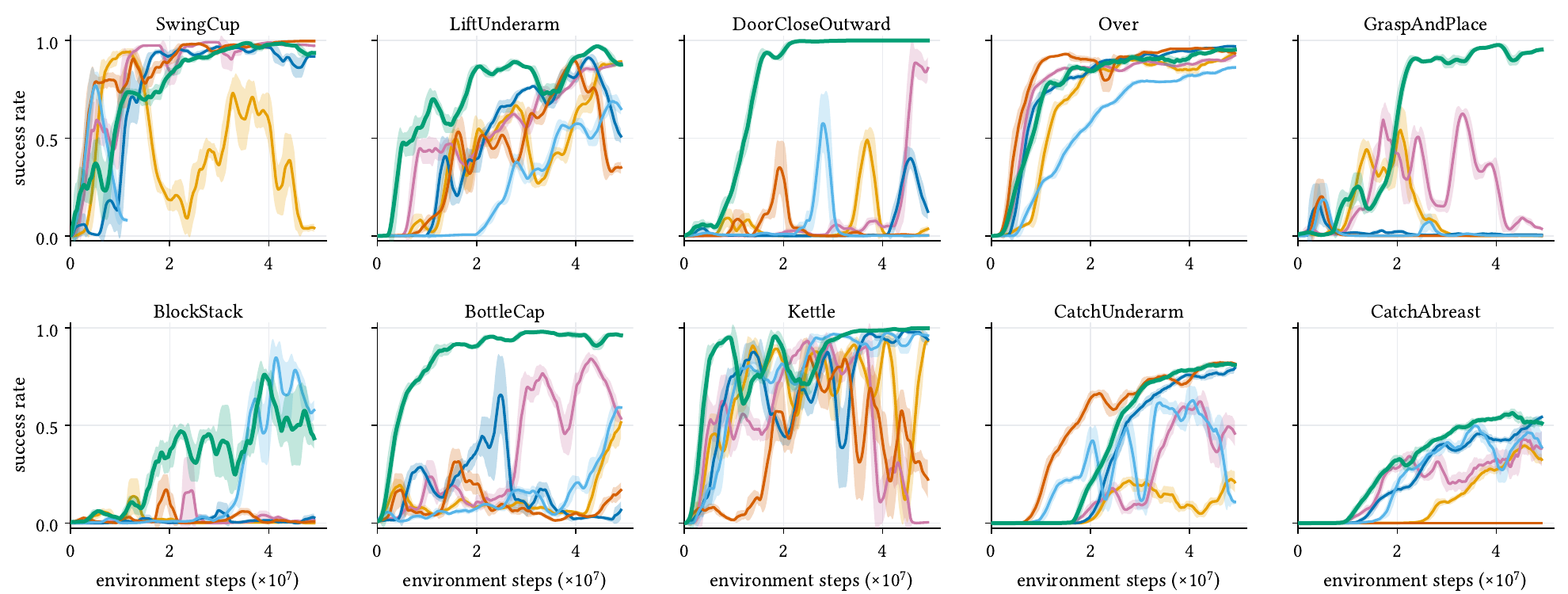}
\vspace{-0.6em}
\caption{Bi-DexHands search and training curves.}
\label{fig:pertask}
\label{fig:bidex-training}
\end{figure}

\Cref{fig:pertask} shows per-task best-so-far scores and training curves. The upper two rows plot the best-so-far success against reward-function samples on the ten Bi-DexHands tasks. The lower two rows plot the success rate of the policy trained with each method's best reward against environment steps. Every method is given the same budget of $128$ reward-function samples.

\subsection{Power-Grid Operation}
\label{sec:app-case14-study}

\begin{figure}[!htbp]
\centering
\includegraphics[width=\textwidth]{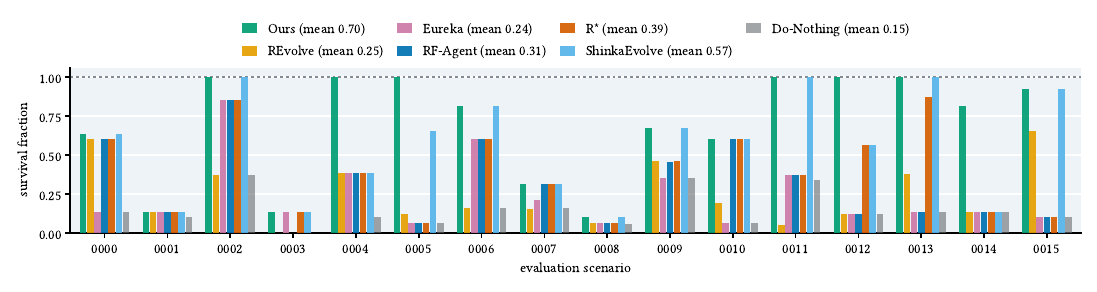}
\caption{Case14 per-scenario survival.}
\label{fig:case14-survival}
\end{figure}

\textbf{Per-scenario survival.} \Cref{fig:case14-survival} replays the official evaluation episodes of each method's best designed reward. Ours completes the most four-week scenarios, ShinkaEvolve completes a few, and no other method completes any. A survival fraction of $1.0$ denotes completion of the full scenario. Reward design has to turn survival into a per-step signal on the loading margin. The control task is sparse, since a scenario is scored only at a blackout or at its end.

\begin{figure}[!htbp]
\centering
\includegraphics[width=\textwidth]{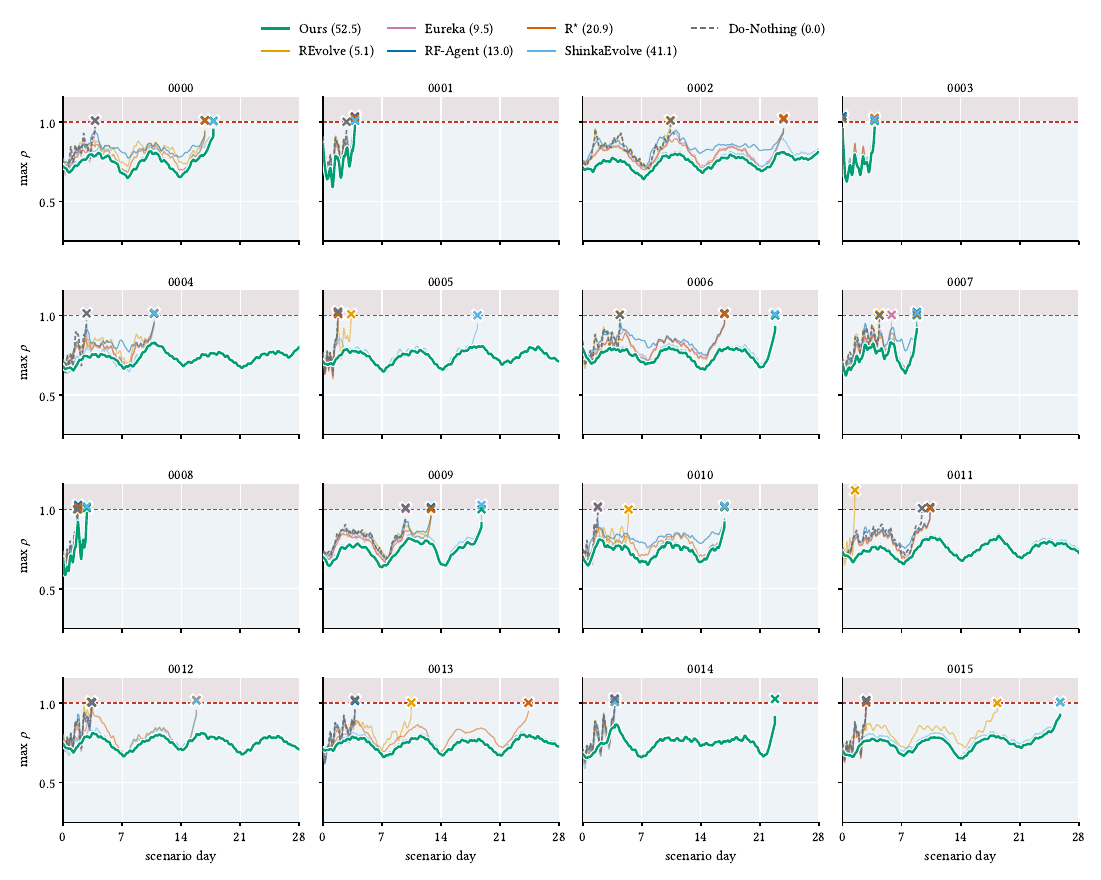}
\caption{Case14 line-loading trajectories.}
\label{fig:case14-agent-study}
\end{figure}

\textbf{Line-loading trajectories.} \Cref{fig:case14-agent-study} shows the maximum line loading $\rho$ during each episode. Each panel is one of the sixteen evaluation scenarios, replayed under each method's best designed reward over four weeks. Dashed curves denote Do-Nothing, and crosses mark blackouts. The red line marks the thermal limit at $\rho=1$, and the legend lists the official score of each method's best reward.

\subsection{Locomotion}
\label{sec:app-loco}

\begin{figure}[!htbp]
\centering
\includegraphics[width=\textwidth]{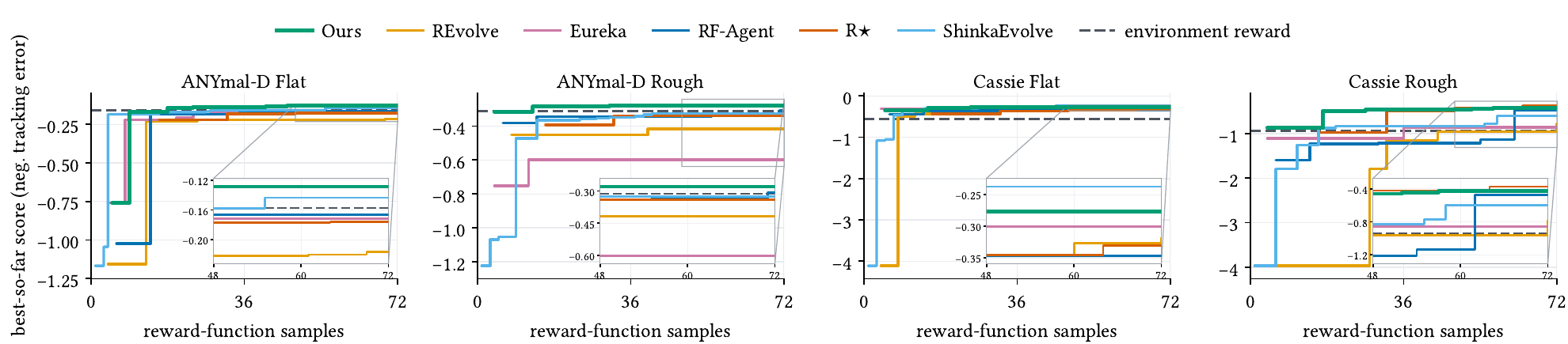}\\[0.7em]
\includegraphics[width=\textwidth]{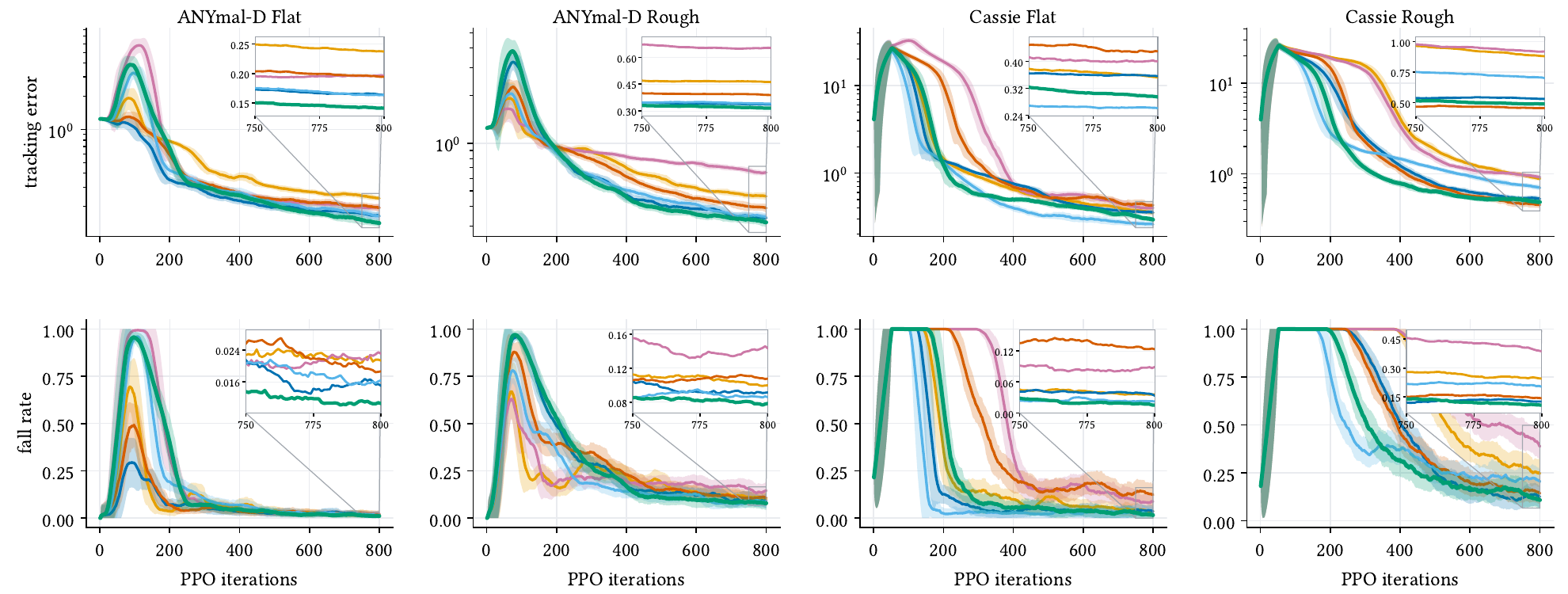}
\caption{Locomotion search and training curves.}
\label{fig:loco}
\end{figure}

\Cref{fig:loco} shows the best-so-far scores during reward search, with performance under the environment reward as a reference, and the PPO training curves for each method's best reward. The panels cover ANYmal-D Flat, ANYmal-D Rough, Cassie Flat, and Cassie Rough, with $72$ reward-function samples per method and $800$ PPO iterations per policy. Search scores are negated fall-charged tracking errors, and the training panels plot the fall-charged tracking error and the fall rate.

\subsection{Traffic-Signal Control}
\label{sec:app-resco}
\begin{figure}[!htbp]
\centering
\includegraphics[width=\textwidth]{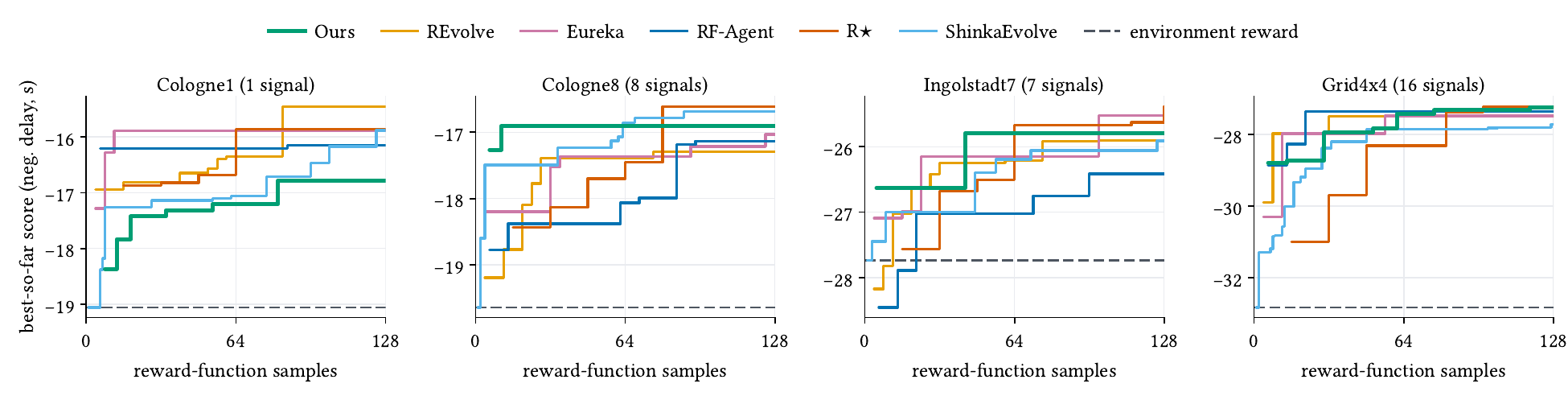}\\[0.7em]
\includegraphics[width=\textwidth]{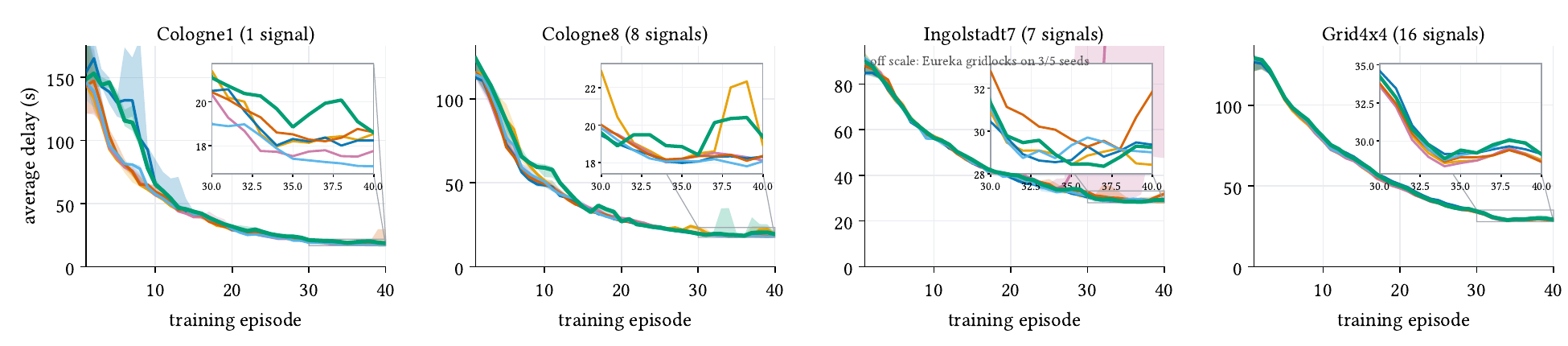}
\caption{RESCO search and training curves.}
\label{fig:resco}
\end{figure}

\Cref{fig:resco} presents the best-so-far scores as a function of the number of evaluated reward candidates and the policy training curves for the reward selected by each method. The panels cover Cologne1, Cologne8, Ingolstadt7, and Grid4x4, with the environment reward as a reference in the search panels. Each method searches for $128$ reward-function samples, and each policy trains for $40$ episodes. The training panels plot the average delay per training episode as the median over five held-out seeds. Bands show the interquartile range across seeds, and insets enlarge the last $10$ episodes.

\subsection{Ablations}
\label{sec:app-ablation-curves}

\begin{figure}[!htbp]
\centering
\includegraphics[width=\textwidth]{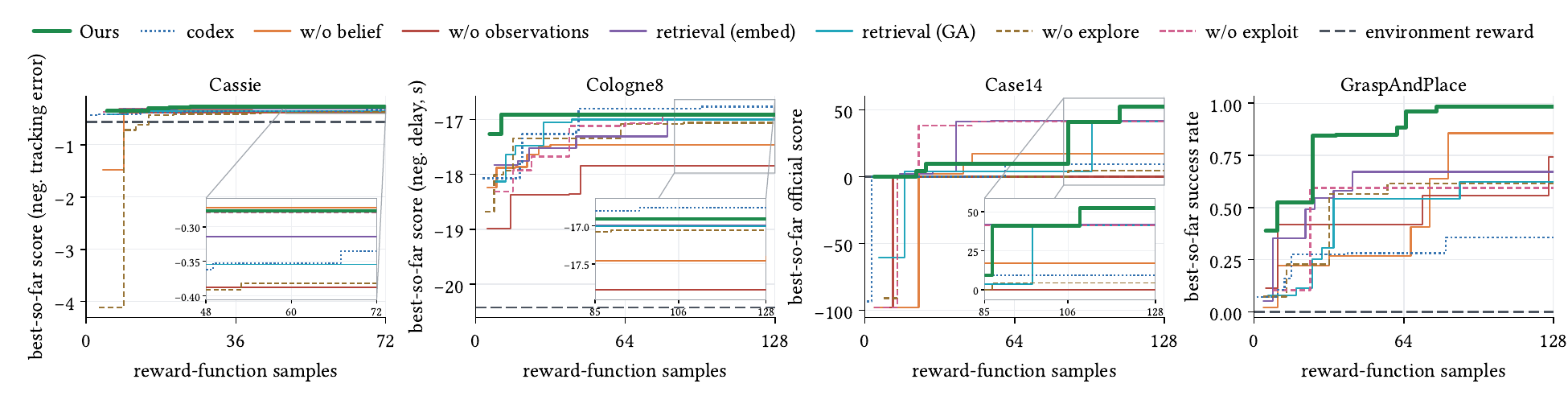}
\caption{History and schedule ablation search curves.}
\label{fig:ablation-curves}
\end{figure}

\textbf{Search curves.} \Cref{fig:ablation-curves} plots the trajectories the bars of \Cref{fig:ablation} end on, each bar the loss as a share of the full method's gain over the environment reward, Do-Nothing on Case14, or Sparse on GraspAndPlace. The four panels show the best-so-far search score against reward-function samples on Cassie Flat, Cologne8, Case14, and GraspAndPlace.

\begin{figure}[!htbp]
\centering
\includegraphics[width=\textwidth]{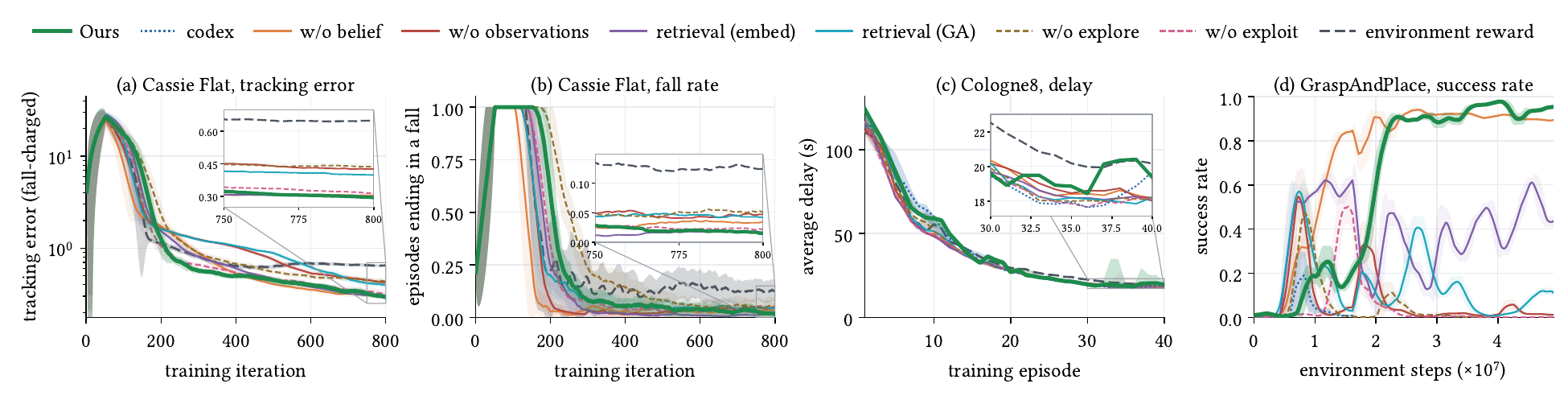}
\caption{History and schedule ablation training curves.}
\label{fig:ablation-training}
\label{fig:ablation-training-cologne8}
\end{figure}

\textbf{Training curves.} \Cref{fig:ablation-training} shows the training curves behind the Cassie Flat, Cologne8, and GraspAndPlace curves of \Cref{fig:ablation-curves} through the environment's own diagnostics. \Cref{fig:ablation-training}a and \Cref{fig:ablation-training}b plot the fall-charged tracking error and the fall rate on Cassie Flat, \Cref{fig:ablation-training}c the average delay on Cologne8, and \Cref{fig:ablation-training}d the success rate on GraspAndPlace.

\begin{figure}[!htbp]
\centering
\includegraphics[width=\textwidth]{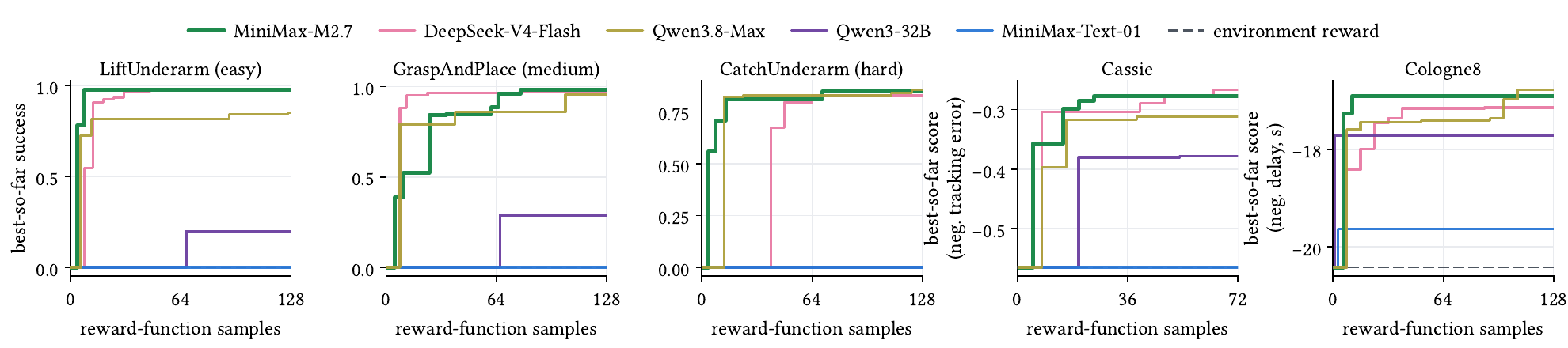}\\[0.7em]
\includegraphics[width=\textwidth]{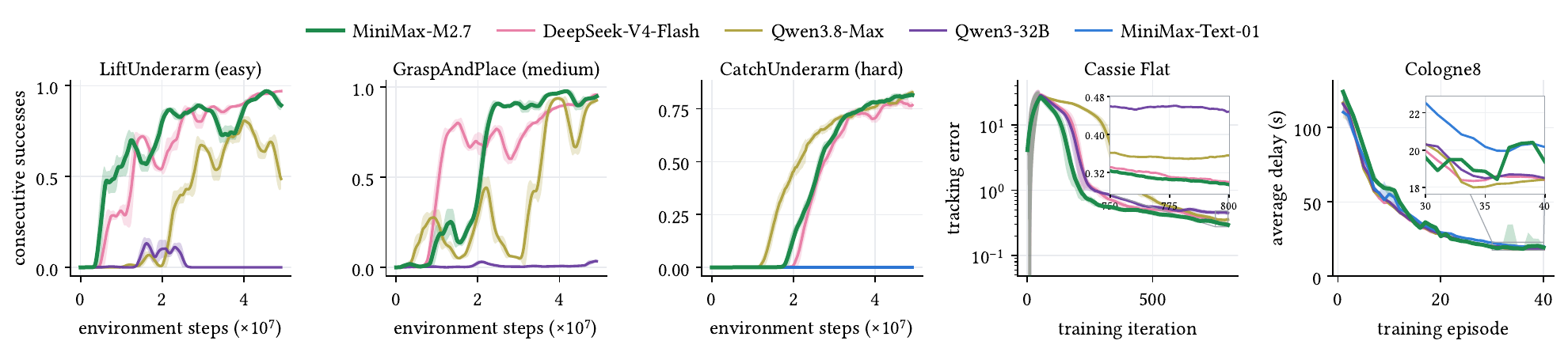}
\caption{Backbone ablation search and training curves.}
\label{fig:backbone-detail}
\label{fig:backbone-training}
\end{figure}

\textbf{Run on different backbones.} \Cref{fig:backbone-detail} presents the best-so-far search scores and policy training curves for each backbone's best reward, using the backbones of \Cref{fig:ablation}c, whose axes run from the reference reward at 0 to the best backbone at 1. MiniMax-Text-01 and Qwen3-32B perform poorly, particularly on manipulation. The method thus needs a capable backbone.

\clearpage
\FloatBarrier
\section{Algorithms}
\label{sec:app-algorithms}
\Cref{alg:search} gives the full procedure of the search described in \Cref{sec:method}.

\begin{algorithm}[H]
\caption{\sys{}: Agentic Reward Black-box Optimization.}
\label{alg:search}
\algrenewcommand{\algorithmiccomment}[1]{\hskip1em{\color{green!45!black}$\triangleright$ \textit{#1}}}
\small\begin{algorithmic}[1]
\Require task specification $\mathcal{T}$, sample budget $B$, schedule constants $n_0,\mu,\nu$
\State $t \gets 1$;\quad $\mathcal{H}_1 \gets \varnothing$;\quad $\mathcal{B}_1 \gets \varnothing$;\quad $\hat R_0 \gets \varnothing$;\quad $F(\varnothing) \gets -\infty$
\While{the sample budget $B$ is not spent}
  \State $n_{\mathrm{rej}} \gets$ consecutive rounds with $a_i=0$ \textbf{and} $m_i=m_{\mathrm{exploit}}$ before $t$
  \State $n_{\mathrm{exp}} \gets$ consecutive rounds after round $n_0$ with $m_i=m_{\mathrm{explore}}$ before $t$
  \If{$t \le n_0$ \textbf{or} $\hat R_{t-1} = \varnothing$} \Comment{schedule $\psi$}
    \State $m_t \gets m_{\mathrm{explore}}$
  \ElsIf{$n_{\mathrm{rej}} \ge \mu$ \textbf{or} $\big(0<n_{\mathrm{exp}}<\nu$ \textbf{and} $a_{t-1}=0\big)$}
    \State $m_t \gets m_{\mathrm{explore}}$
  \Else
    \State $m_t \gets m_{\mathrm{exploit}}$
  \EndIf
  \State $\mathcal{D}_t \gets \mathcal{H}_t \cup \mathcal{B}_t$ \Comment{history}
  \State $\{R_t^k\}_{k=1}^{N_t} \sim g\big(\cdot\mid\mathcal{T},\mathcal{D}_t,m_t\big)$ \Comment{proposal operator}
  \State $\mathcal{B}_{t+1} \gets G_{t+1} \cup \Lambda_{t+1}$
  \For{$k=1,\dots,N_t$}
    \State $(y_t^k,\ell_t^k,e_t^k) \gets \mathcal{O}(R_t^k)$ \Comment{evaluation oracle}
  \EndFor
  \State $\mathcal{H}_{t+1} \gets \mathcal{H}_t \cup \{(R_t^k,y_t^k,\ell_t^k,e_t^k)\}_{k=1}^{N_t}$
  \State $k^\star \gets \arg\max_k y_t^k$;\quad $a_t \gets \mathbf{1}\big[y_t^{k^\star} > F(\hat R_{t-1})\big]$
  \State $\hat R_t \gets R_t^{k^\star}$ \textbf{if} $a_t=1$ \textbf{else} $\hat R_{t-1}$
  \State $t \gets t+1$
\EndWhile
\State \Return the best reward $\hat R_{t-1}$, the highest-scoring candidate in $\mathcal{H}_t$
\end{algorithmic}
\end{algorithm}

\section{Licenses}
\label{sec:app-licenses}
\Cref{tab:app-licenses} lists the licenses and URLs of the baselines, benchmarks and training algorithms we use in this work. R$\star$ releases no code, and our reimplementation follows Appendix~\ref{sec:app-baselines}.

\begin{table}[H]
\centering
\caption{Resource licenses and URLs.}
\label{tab:app-licenses}
\scriptsize
\renewcommand{\arraystretch}{1.0}
\setlength{\tabcolsep}{5pt}
\resizebox{0.97\textwidth}{!}{%
\begin{tabular}{@{}lll@{}}
\toprule
Resources & License & URL \\
\midrule
Eureka~\citep{eureka} & MIT & \url{https://github.com/eureka-research/Eureka} \\
REvolve~\citep{revolve} & Available online & \url{https://github.com/RishiHazra/Revolve} \\
RF-Agent~\citep{rfagent} & Available online & \url{https://github.com/deng-ai-lab/RF-Agent} \\
ShinkaEvolve~\citep{shinka} & Apache 2.0 & \url{https://github.com/SakanaAI/ShinkaEvolve} \\
\midrule
Isaac Gym~\citep{isaacgym} & BSD 3-Clause & \url{https://github.com/isaac-sim/IsaacGymEnvs} \\
Bi-DexHands~\citep{bidexhands} & Apache 2.0 & \url{https://github.com/PKU-MARL/DexterousHands} \\
Grid2Op~\citep{grid2op} & MPL 2.0 & \url{https://github.com/Grid2Op/grid2op} \\
Isaac Lab~\citep{isaaclab} & BSD 3-Clause & \url{https://github.com/isaac-sim/IsaacLab} \\
RESCO~\citep{resco} & CC BY-NC-SA 4.0 & \url{https://github.com/Pi-Star-Lab/RESCO} \\
SUMO~\citep{sumo} & EPL 2.0 & \url{https://github.com/eclipse-sumo/sumo} \\
\midrule
rl-games~\citep{rlgames} & MIT & \url{https://github.com/Denys88/rl_games} \\
Stable-Baselines3~\citep{sb3} & MIT & \url{https://github.com/DLR-RM/stable-baselines3} \\
RSL-RL~\citep{rslrl} & BSD 3-Clause & \url{https://github.com/leggedrobotics/rsl_rl} \\
PFRL & MIT & \url{https://github.com/pfnet/pfrl} \\
\bottomrule
\end{tabular}}
\end{table}

\clearpage
\section{Prompts}
\label{sec:app-skill}
The fixed context $c_t$ of \Cref{eq:operator} consists of the texts shown below. The \textbf{instruction file} is the fixed instruction of the proposal operator. The \textbf{search mode guidance} is the steering prompt $m_t$ of \Cref{eq:schedule}, and its two values differ only in the mode named beneath the protocol text. The \textbf{task prompt} carries the task specification $\mathcal{T}$, the task description whose source Appendix~\ref{sec:app-environments} cites, with its domain reward contract and the acceptance metric $F$. 

\textbf{Instruction file.} The instruction file states what the workspace holds and what one round must produce, and it leaves the task specification to the slots highlighted in blue.
\begin{promptbox}{Instruction file: the per-round workflow.}
\begin{lstlisting}[style=rfprompt]
---
name: reward_design
scope: subagent
agent: reward_design
description: >-
  Design and improve the training reward for a reinforcement-learning task. Read the staged task and the history of past attempts, write this attempt's candidate rewards, and commit them once for evaluation.
initial_message: >-
  Improve the training reward for this reinforcement-learning task:

  {{ task_specific_prompt }}

  The task prompt above is authoritative for this task, and it names the reward contract, the files you write, and the acceptance metric the loop maximizes.

  `.arbo_history/observations.jsonl` in the staged code package is the record of every past attempt and of every candidate inside it, and `analysis.md` holds your own notes from the last one. An absent or empty history means this is the first attempt.

  You have {{ max_turns }} tool-calling turns, and the attempt scores nothing until candidates are written and committed, so make your first write by turn {{ write_by_turn }}.

  Each attempt you read what you need, write this round's candidates, overwrite `analysis.md` with your evidence and one hypothesis per candidate, and commit them in exactly one commit. Then stop and let the host evaluate.

  {{ search_mode_guidance }}

  search_mode: {{ search_mode }}
---

# Reward design

Your reward is a training signal. The acceptance metric is fixed, it is defined by the task, and it is never yours to edit or reimplement. Your reward rising while the acceptance metric does not is the failure this loop selects against.

The task file in the staged code package, named in the task prompt, is the complete task. It holds the state your reward may consume, with names, shapes, frames and units, and it holds the code that computes the acceptance metric. Nothing outside the files the task prompt names is yours to edit.

`observations.jsonl` records every candidate and not only the winners, so each one carries its score, its per-term diagnostics, and its traceback if it crashed. A crashed candidate still spent a sample and says nothing about the hypothesis behind it.

Training is deterministic, and a candidate whose code repeats an earlier one is not retrained but still costs a sample.
\end{lstlisting}
\end{promptbox}

\textbf{Search mode guidance.} The search mode guidance is the static explore/exploit protocol, injected into the context each round by the schedule.
\begin{promptbox}{Search mode guidance: explore/exploit protocol.}
\begin{lstlisting}[style=rfprompt]
The host sets your mode each attempt and scores the result, and it never runs git, so you run every git command yourself inside the staged `code_package`. Every attempt ends with exactly one commit on a named branch, never on a detached HEAD, where `main` carries the incumbent best line and each explore probe lives on its own `explore/*` branch.

In exploit mode you refine the current best, keeping the method and tuning it rather than discarding what works, so run `git checkout main` first, and if your last winning probe is still on an `explore/*` branch, fold it in with `git merge --no-ff <branch>` before you commit. In explore mode you propose a structurally different candidate, a different method rather than different constants, so run `git checkout -b explore/<short-name> <bootstrap-sha>` and branch off the bootstrap root rather than `main` or an earlier probe. Either way the host accepts or rejects by score alone, independent of which branch you chose.
\end{lstlisting}
\end{promptbox}

\textbf{Task prompt.} The task prompt is rendered from the task adapter when the workspace is staged. The text below shows its slots in double braces.
\begin{promptbox}{Task prompt: the per-task slot.}
\begin{lstlisting}[style=rfprompt]
Task id: {{ task_id }}
{{ task_description }}

Write N candidate rewards this attempt, N in [1, {{ N_max }}]; files beyond the cap are dropped. The host trains all N in parallel and every file costs one sample. This run must reach {{ max_samples }} evaluated samples within at most {{ max_attempts }} attempts; observations.jsonl shows how many of each are used.
Put them in this attempt's candidate folder, named in your instructions, as reward1.py .. rewardN.py together with this attempt's analysis.md, and commit them in one commit. Only that folder is trained; earlier iteration_* folders are read-only history.

Reward contract: {{ reward_contract }}

shared/input_assets/code_package/env.py is the full composed task and is read-only. Every candidate trains for the same fixed budget.

The loop steers explore/exploit by the best candidate's `{{ acceptance_key }}` — this task's acceptance metric ({{ metric_kind }}; {{ metric_range }}). The task description above states the control objective; read the top-level env.py documentation and the code that computes `{{ acceptance_key }}` for its exact evaluation formula. Design each candidate as a TRAINING reward that teaches the policy to achieve that objective; never replace or imitate the fixed evaluator mechanically. The metric may be negative or unbounded. Raise it (higher is always better), and do not assume a [0,1] range.

Metrics in observations.jsonl: `{{ acceptance_key }}_*` is the acceptance metric ({{ metric_kind }}; {{ metric_range }}). The reported score is the {{ aggregation }} of `{{ acceptance_key }}`; the loop accepts an attempt ONLY if it rises. Judge by TREND, not absolute value — do not assume a [0,1] range. `gpt_reward_*` is the reward the policy actually optimized (your designed reward's realized value) — its scale/rise is not by itself task progress. `rew_<term>_*` are your per-term diagnostics revealing dead, saturated, exploding, or dominating components.

Each line of .arbo_history/observations.jsonl is one attempt: `mode`, `decision`, `reward` (its acceptance value, `{{ acceptance_key }}`), and `raw.candidates[]` with one record per candidate (`file`, `score`, `metrics`, `series`, `error`, `train_log`). The lines are large; project the fields you want with python or jq.
\end{lstlisting}
\end{promptbox}

\textbf{AutoResearch with Codex.} \emph{codex} in \Cref{fig:ablation}b receives the prompt below.
\begin{promptbox}{AutoResearch with Codex: the session prompt.}
\begin{lstlisting}[style=rfprompt]
Design the training reward for a reinforcement-learning control task.

## Task: {{ task_id }}

{{ task_description }}

## Files

`reward.py` holds the reward function. It starts as a placeholder that returns
zero everywhere and trains nothing -- a valid starting point, not a reward.

`env_reference.py` is the complete environment code, read-only. It defines the
state tensors a reward may read, the episode structure, the resets, and the
success criterion. Read it to learn which `self.<name>` tensors exist and what
they mean.

## Reward contract

{{ reward_contract }}

## Scoring a reward

    evaluate <file.py> [<file.py> ...]

Each file is one candidate: the command trains a policy on it with {{ rl_algorithm }}
from scratch and scores the resulting policy by the task's own acceptance metric. The
metric is fixed and you cannot edit or influence it except through the reward you write:

  * `{{ acceptance_key }}` -- {{ metric_kind }}
  * range: {{ metric_range }}
  * aggregation: {{ aggregation }}

{{ official_evaluator_note, when the task has one }}

For each candidate it reports the score, the training statistics of your reward's own
named terms, the tail of the training log, and the full traceback if it failed. It is
the only way to find out how a reward performs -- there is no network here and no way
to run the environment yourself.

The command blocks: training is real and takes roughly {{ eval_minutes }} minutes per
candidate. Candidates passed in one call train in parallel, so a call with several
candidates costs about as much wall-clock as a call with one.

## Budget

You have {{ n_budget }} evaluations for this task. Every file you pass to `evaluate` spends
one, whether or not the reward turns out to work. Nothing else is measured, and there
is nothing beyond the budget: when it is spent, `evaluate` refuses.

Your goal is the highest `{{ acceptance_key }}` that any single candidate of yours
reaches within those {{ n_budget }} evaluations. How you get there is up to you.
\end{lstlisting}
\end{promptbox}

\end{document}